\documentclass[acmlarge]{acmart}

\AtBeginDocument{%
  \providecommand\BibTeX{{%
    \normalfont B\kern-0.5em{\scshape i\kern-0.25em b}\kern-0.8em\TeX}}}

\setcopyright{acmcopyright}
\acmJournal{JETC}
\acmYear{2020} 
\acmVolume{1} 
\acmNumber{1} 
\acmArticle{1} 
\acmMonth{1} 
\acmPrice{15.00}
\acmDOI{10.1145/3399670}

\usepackage[normalem]{ulem}
\usepackage{url}
\usepackage{hyperref}
\usepackage{color}
\usepackage{soul}
\usepackage{mathptmx} 
\usepackage{microtype}
\usepackage{graphicx,wrapfig,lipsum}
\usepackage{tabularx}
\usepackage{fancyhdr}
\usepackage{geometry}
\usepackage{flushend}
\usepackage{multirow}
\usepackage{arydshln}
\usepackage{color}
\usepackage{subfig}
\usepackage{textcomp}
\usepackage{tikz}
\usepackage{comment}
\usepackage[font=small,skip=2pt]{caption}
\usepackage{makecell}

\newcommand*\circled[1]{\tikz[baseline=(char.base)]{\node[fill=white,shape=circle,draw,inner sep=0.6pt] (char) {#1};}}
\begin{document}

\title{RNNFast: An Accelerator for Recurrent Neural Networks Using Domain Wall Memory}

\author{Mohammad Hossein Samavatian}
\affiliation{%
  \institution{The Ohio State University}
  \streetaddress{}
  \city{Columbus}
  \state{Ohio}
  \country{USA}}
\email{samavatian.1@osu.edu}

\author{Anys Bacha}
\affiliation{%
  \institution{University of Michigan}
  \streetaddress{}
  \city{Dearborn}
  \state{Michigan}
  \country{USA}}
\email{bacha@umich.edu}

\author{Li Zhou}
\affiliation{%
  \institution{The Ohio State University}
  \streetaddress{}
  \city{Columbus}
  \state{Ohio}
  \country{USA}}
\email{zhou.785@osu.edu}

\author{Radu Teodorescu}
\affiliation{%
  \institution{The Ohio State University}
  \streetaddress{}
  \city{Columbus}
  \state{Ohio}
  \country{USA}}
\email{teodorescu.1@osu.edu}
\begin{abstract}

Recurrent Neural Networks (RNNs) are an important class of neural networks designed to retain and incorporate context into current decisions. RNNs are particularly well suited for machine learning problems in which context is important, such as speech recognition and language translation. 

This work presents RNNFast, a hardware accelerator for RNNs that leverages an emerging class of non-volatile memory called 
domain-wall memory (DWM). We show that DWM is very well suited for RNN acceleration due to its very high density and low read/write energy. At the same time, the sequential nature of input/weight processing of RNNs mitigates one of the downsides of DWM, which is the linear (rather than constant) data access time. 

RNNFast is very efficient and highly scalable, with flexible mapping of logical neurons to RNN hardware blocks. The basic hardware primitive, the RNN processing element (PE) includes custom DWM-based multiplication, sigmoid and tanh units for high density and low-energy. The accelerator is designed to minimize data movement by closely interleaving DWM storage and computation. We compare our design with a state-of-the-art GPGPU and find $21.8\times$ higher performance with $70\times$ lower energy. 

\end{abstract}



\renewcommand{\textuparrow}{$\uparrow$}

\maketitle
\section{Introduction}
\label{sec:introduction}

Deep learning is transforming the way we approach everyday
computing. From speech recognition that empowers today's digital
assistants to business intelligence applications
fueled by the analysis of social media postings,
processing information in a way that preserves
the correct context is crucial. For instance, the sentences
``white blood cells destroying an infection'' and
``an infection destroying white blood cells'' have
very different meanings even though they contain the
same words. Traditional machine learning designs such as Convolutional Neural
Networks (CNNs) do not consider context and are therefore not well suited
for solving such problems.
Recurrent Neural Networks (RNNs) are a powerful class of networks designed to consider context by retaining and using information from previously processed inputs. RNNs are used across a wide range of applications that include speech recognition for digital assistants such as Siri and Google Now, sentiment analysis
for classifying social media postings, and language translation. The popularity of RNN networks in production applications was highlighted by Google in a recent paper \cite{TPU_isca_17}, which reports that RNN workloads represent almost 30\% of the workloads on Google's TPU datacenters. This is in contrast to only 5\% for CNN workloads. 

However, RNN workloads are data-intensive because they store a partial history of the output sequence and perform computations on that history along with the current input. As a result, RNNs require both vast amounts of storage and increased processing power. For example, the RNN neuron requires 8$\times$ the number of weights and multiply-accumulate (MAC) operations of a typical CNN cell. RNN networks are also generally quite large. For instance, Amodei et al. \cite{amodei_icml2016} developed a network for performing speech recognition that utilized seven recurrent layers and a total of 35 million parameters. At this scale, RNNs with large input sets are susceptible to memory bottlenecks when running on existing accelerators such as GPUs \cite{Guan_ASP_DAC_17} or FPGAs \cite{Guan_ASP_DAC_17, Li_FPCCM_15, Ferreira_16ReConfig,mealey2018accelerating,Azari-FPGA19,Warg-JETCAS19,li2018rnn,Wang-FPGA18}. In addition, the fundamentally different design of the RNN cell makes previously proposed custom CNN accelerators \cite{Venkataramani_ISCA17,Shen_ISCA17,Parashar_ISCA17,chen_micro2014,chen_asplos2014, liu_asplos2015, du_isca2015, han_isca2016, chen_isca2016,chang-ISCAS19, 
shafiee_isca2016, chi_isca2016, kim_isca2016, albericio_isca2016, xwang-HPCA19,Jin-ASPLOS19, Kung-ASPLOS19,
likamwa_isca2016, reagenisca2016, liu_isca2016} not directly applicable to RNN workloads.

This paper presents RNNFast, a hardware accelerator for RNN networks. RNNFast leverages domain-wall memory (DWM), an emerging non-volatile memory technology, to provide high density on-chip storage as well as energy efficient computation. DWM \cite{Parkin190,wang_nanotech2015, wang_date2014, yu_aspdac2014, huang_vlsi2016,chung_islped2016, zhao_ftfc2013}  is a
magnetic spin-based memory technology, which stores information by
setting the spin orientation of so-called magnetic domains in a
ferromagnetic wire. Multiple magnetic domains can occupy a single wire
(referred to as ``racetrack'') allowing up to 64 bits to be
represented.

DWM has many attractive characteristics. It has read/write latencies that are close to SRAM and write performance and energy that are substantially lower than STT-RAM and other non-volatile memories \cite{Venkatesan_2016}. Perhaps more importantly, DWM is expected to have ~30$\times$ higher density than SRAM and ~10$\times$ higher than DRAM or STT-RAM. The technology would therefore allow dramatically higher storage capacity in the same chip area. While the technology is still in the early stages of development, prototypes have yielded encouraging results \cite{Annunziata_11}. We show that DWM is very well suited for RNN acceleration due to its very high density, linear access pattern, and low read/write energy.

The RNNFast architecture is  modular and highly scalable forgoing the need for long communication buses despite the high output fanout of typical RNN networks. RNNFast allows flexible mapping of logic neurons to RNN hardware blocks. The accelerator is designed to minimize data movement by closely interleaving DWM storage and computation. The basic hardware primitive, the RNN processing element (PE) includes custom DWM-based multiplication and custom nonlinear functional units for high performance and low-energy. RNNFast also includes an error mitigation mechanism for position errors, expected to be relatively common in DWM. The error mitigation is tailored to the RNNFast data access pattern to minimize overhead. We compare RNNFast with a state-of-the art  NVIDIA P100 GPGPU and find RNNFast improves performance by $21.8\times$ while reducing energy $70\times$.

We also compare with two alternative RNNFast designs. 
1) a CMOS-based RNNFast design in which both memories and logic use traditional CMOS. We find the RNNFast design to be up to 2$\times$ more energy efficient than the CMOS version, in a much smaller chip area. 2) a memristor-based implementation that uses an analog dot-product engine, a state-of-the-art design that has been shown to be very efficient for CNNs \cite{chi_isca2016,ankit2017resparc}. RNNFast shows better performance, energy and area than the memristor-based design.
Qualitative comparisons with FPGA-based RNN accelerators, Google's TPU and Microsoft's Brainwave \cite{brainwave_isca18} also indicate RNNFast has better performance and lower energy for similar workloads. 

This paper makes the following main contributions:
\begin{itemize}
\item Presents RNNFast, the first DWM-based custom accelerator for LSTMs and other RNN variants.
\item Introduces novel DWM-based designs for efficient neural network hardware including sigmoid, and tanh  units.
\item Implements an efficient error mitigation solution for DWM overshift errors.
\item Presents a new efficient and scalable interconnection mechanism based on racetrack chains. 
\item Demonstrates that DWM is very well suited for efficient acceleration of recurrent neural networks.
\end{itemize}

The rest of this paper is organized as follows:
Section \ref{sec:background} provides background
information. Section \ref{sec:architecture} details the design and
implementation of RNNFast. Section \ref{sec:errors} presents the error mitigation aspects of the design. Sections \ref{sec:method} and \ref{sec:evaluation}
describe the evaluation.
Section \ref{sec:related} discusses related
work and Section \ref{sec:conclusions} concludes.

\section{Background}
\label{sec:background}
 Recurrent neural networks (RNN) are a powerful class of networks that
 have the ability to learn sequences.  They are applicable
 to anything with a sense of order that needs to be remembered.
 RNNs are used across a wide range of applications that
 includes speech recognition for enabling today's digital
 assistants, sentiment analysis for analyzing posts
 (text and video) and classifying them as positive
 or negative, and machine translation for sequence
 to sequence translation between languages.

\begin{figure}
  \centering
     \includegraphics[width=0.5\linewidth]{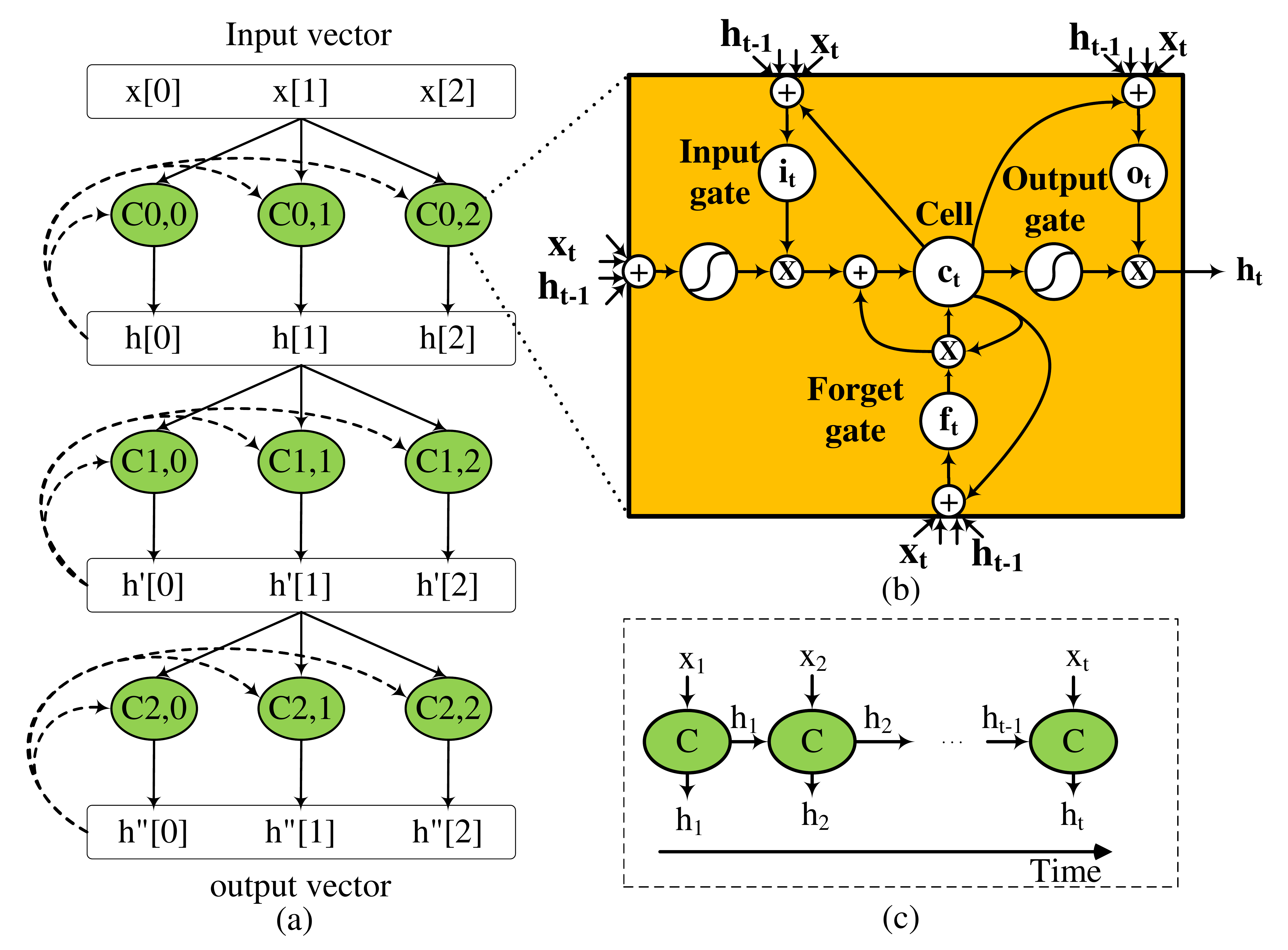}
  \caption{\small (a) 3-layer RNN with 3 LSTM cells/layer, (b) LSTM cell, (c) an LSTM cell unrolled over time}\label{Fig.:LSTM}
  \vspace{-1em}
\end{figure}

\subsection{The Long Short-Term Memory Cell}


Most recurrent neural networks make use of special "neurons"
called Long Short-Term Memory (LSTM) cells \cite{hochreiter_neco97,greff2016lstm}.
LSTMs are designed to process and remember prior inputs and factor them into their outputs over time. Figure \ref{Fig.:LSTM} shows an example of a very simple 3-layer RNN with 3 LSTM cells/layer. The output of each layer is a vector that is supplied as the input to the following layer. In addition to those inputs, a feedback loop takes the output vector of each layer and feeds it back as an additional input to each LSTM neuron. An illustration of the inputs and outputs of a single LSTM cell $C$ unrolled over time is shown in Figure \ref{Fig.:LSTM}(c). An input $x_0$ into neuron $C$ at time step $t = 0$,
will generate an output $h_0$ that is propagated downstream
to the next layer. In addition, $h_0$ is saved within
the neuron's memory cell for use in the next time step.
At time step $t = 1$, the same
neuron $C$ will process input $x_1$, but also use
the previously stored output $h_0$ to generate
the new output $h_1$.

A detailed look inside the LSTM neuron (Figure \ref{Fig.:LSTM}(b)) reveals a significantly more complex operation compared to CNN neurons. The strength of the LSTM lies in the way it regulates the fraction of information it recalls from its embedded memory and the fraction of input it processes for generating outputs over time. In other words, the LSTM cell progressively memorizes and forgets contextual information as it processes more inputs.
This is achieved through special gates that are controlled
through a set of mathematical functions \cite{graves_icassp13} governed by equations \eqref{it_eq} -- \eqref{ht_eq}.
\begin{equation} \label{it_eq}
 i_t = \sigma(W_{xi}x_t + W_{hi}h_{t-1} + b_i)
\end{equation}
\begin{equation} \label{ft_eq}
 f_t = \sigma(W_{xf}x_t + W_{hf}h_{t-1} + b_f)
\end{equation}
\begin{equation} \label{ot_eq}
 o_t = \sigma(W_{xo}x_t + W_{ho}h_{t-1} + b_o)
\end{equation}
\begin{equation} \label{ct_eq}
 c_t = f_t \odot	c_{t-1} + i_t \odot tanh(W_{xc}x_t + W_{hc}h_{t-1} + b_c)
\end{equation}
\begin{equation} \label{ht_eq}
 h_t =o_t \odot tanh(c_t)
\end{equation}
The input gate $i_t$ receives the input to be written into
a neuron's memory cell at time step $t$. The forget gate
$f_t$ controls what information should be erased from a
neuron's memory cell at time step $t$. The cell $c_t$
represents the content of the neuron's memory cell. The output gate $o_t$ controls the
amount of information read from the neuron's cell and
how much of it contributes to the output.
The output $h_t$ represents the output of the cell to the
next layer at time step $t$. This output is also fed back
into the input gate $i_{t+1}$ of the same LSTM cell at time
step $t+1$. The $W$s and $b$s represent the weights and biases, respectively.

Note that $\odot$ used in equations \eqref{ct_eq} and \eqref{ht_eq} represents
the dot product operator. In addition, equations \eqref{it_eq} -- \eqref{ht_eq}
represent neurons for an entire layer within a network. Therefore,
$i_t$, $f_t$ , $o_t$, $c_t$, $h_t$, $h_{t-1}$, and $x_t$
are vectors and all $W$s are matrices.
As such, if we augment a given matrix $W$ to include
the weights for both $x$ and $h$ such that its dimensions
are $n \times m$, then each row in $W^l$
for hidden layer $l$ would be mapped to neuron $j$
where $j \in [1,n]$. The value m is the size of input vector.

\begin{equation}  \label{weight_eq}
\small  W^l = \begin{bmatrix}W^l_{11} & ... & W^l_{1m} \\ \vdots & \ddots & \vdots \\ W^l_{n1} & ... & W^l_{nm} \end{bmatrix}
\end{equation}
The $tanh$ and $\sigma$ activation functions are also outlined
in equations \eqref{sig_eq} and \eqref{tanh_eq} for clarity.
These functions are applied as elementwise operations on
the resulting vectors.
\begin{equation} \label{sig_eq}
\sigma\left(z\right) = \frac{1}{1 + e^{-z}}
\end{equation}
\begin{equation}  \label{tanh_eq}
tanh\left(z\right) = 2\sigma \left( 2z \right) -1
\end{equation}

\begin{figure}
  \centering
     \includegraphics[width=0.7\linewidth]{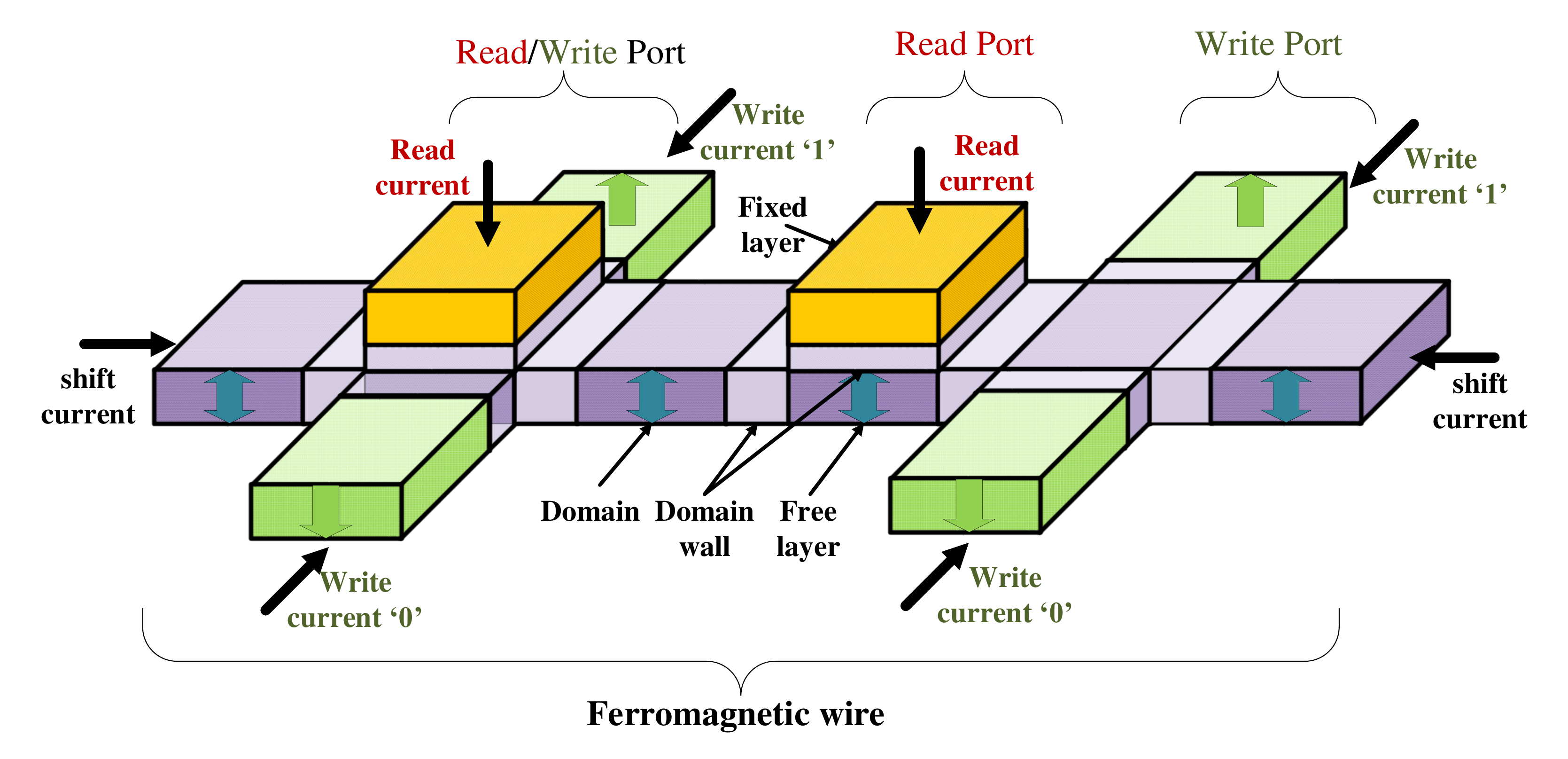}
  \caption{DWM device structure.}\label{Fig.:DWM}
  \vspace{-1em}
\end{figure}

Because of the complex design, LSTM cells require substantially more storage and computation relative to their CNN counterparts. Moreover, RNN networks are also generally fully-connected, further increasing the data movement overhead. 

\subsection{Domain-wall Memory}
\label{sec:background_dwm}

Domain wall (a.k.a. racetrack) memory was first proposed by Parkin et al. \cite{Parkin190} from IBM in 2008. In 2011, Annunziata et al.\cite{Annunziata_11} demonstrated the first 200mm DWM wafer, fabricated with IBM 90nm CMOS technology. Each die contained 256 racetrack cells, proving the feasibility of DWM fabrication. A large body of research has since sought to improve and optimize the technology at device and circuit levels \cite{Sun:2013:CRM:2463209.2488799,Ghosh_DAC13,Sun_TC16,Zhang_TCS16,Motaman_tn15,Zhang_aspdac15,Venkatesan_date13} and find solutions to improve its reliability \cite{Zhang_ISCA_15}.

Domain wall (racetrack) memory represents information using the spin orientation of magnetic domains in a ferromagnetic wire, as shown in Figure~\ref{Fig.:DWM}. Each of these domains can be independently set to an up-spin or down-spin to represent the value of a single bit. Since multiple magnetic domains can reside on a single wire, multiple bits (32-64) of data can be packed in a single DWM device, resulting in a very high density. Three basic operations can be performed on a DWM device: read, write and shift. A magnetic tunnel junction (MTJ) \cite{xuiccad11, smulleniccad11} structure is used to read data from the DWM cell (read port in Figure~\ref{Fig.:DWM}).

In a DWM device, all the magnetic domains share a single read MTJ (generally referred-to as a read head or port). The bit to be read needs to be aligned with the MTJ before it can be accessed. This is accomplished using a property that is unique to DWM, called domain wall motion, which refers to the shifting of magnetic domains down the ferromagnetic wire. When a current pulse of a suitable magnitude is applied through the ferromagnetic wire, the magnetic spins of all domains ``move'' across the wire in a direction opposite to the direction of the current. The number of bit positions in a shift motion is controlled by the duration of the shift current. Additional blank domains are included at the ends of each racetrack to allow all data domains to be shifted to the read head without data loss at the ends of the wire \cite{Ranjan_DATE_15}.  

Writing into DWM is also fast and energy efficient due to recently developed \cite{Zhang_aspdac15} "shift-based writes" as demonstrated in Fig.~\ref{Fig.:DWM} (write port). The design of the write head consists of a ferromagnetic wire with two fixed domains that straddle a free domain at an arbitrary location on the racetrack. One of the fixed domains is hardwired to up-spin and the other to down-spin at fabrication. The spin of either of the fixed domains can be shifted into the free domain through the domain motion process by applying a current pulse in the appropriate direction. The latency and energy of shift-based writes are equivalent to those of simple shifts. 



The main challenge of racetrack memory is the access latency to data stored in a DWM tape which is variable depending upon the number of shifts required to align the accessed bit with the read or write heads. RNNFast mitigates this disadvantage by optimizing data placement for sequential access such that most accesses only require a single shift.

\subsubsection{Reliability Issues}

DWM technology also presents reliability challenges including possible misalignment of the data domains leading to erroneous reads and/or writes \cite{Iyengar_DAC_14,Zhang_ISCA_15}. Prior work \cite{Zhang_ISCA_15} has classified DWM errors into two main types: "stop-in-the-middle" and "out-of-step" errors. The first class of errors is caused when data domains are not aligned with the read/write heads, leading to invalid accesses. The second class of errors is caused when the incorrect domain is aligned with the read/write head which causes the wrong bit in the track to be accessed. 
The errors are generally caused by variability in the magnitude or duration of the current pulse applied during the domain shift operation. Zhang et al.\cite{Zhang_ISCA_15} has developed a technique for eliminating "stop-in-the-middle" errors that relies on the application of a short subthreshold shift current to nudge the misaligned domain back into alignment. They also demonstrate that the subthreshold pulse is small enough that it cannot misalign a correctly aligned domain. As a result, sub-threshold shifts can virtually eliminate "stop-in-the-middle" errors, at the cost of increasing the number of "out-of-step" errors. 

While subthreshold shifts can be applied in both directions, we choose to apply them in the shift direction. As a result, all "out-of-step" errors will be converted into overshift errors by 1 or more positions in the shift direction. For a single-position shift, which represents virtually all shifts in RNNFast,  the probability of single-bit overshift is on the order of $10^{-5}$ \cite{Zhang_ISCA_15}, which is quite high. However, the probability of multibit overshift is about $10^{-21}$, which is negligible. As a result, RNNFast implements mitigation for single-bit overshift errors.

\begin{figure}
\centering
\includegraphics[width=0.6\linewidth,trim = 0 0 0 0]{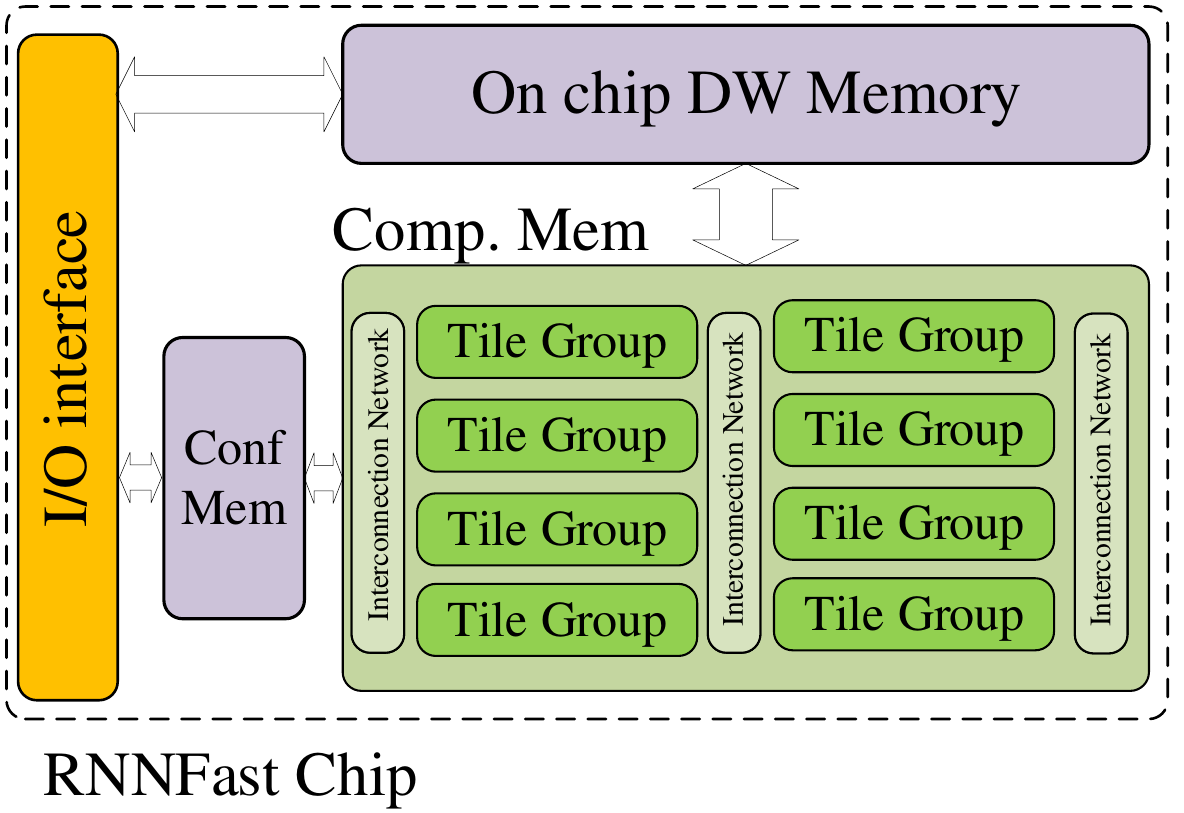}
\caption{RNNFast architecture overview at chip level.}\label{Fig.:architecture}
\vspace{-1em}
\end{figure}
\section{RNNFast Architecture}

\label{sec:architecture}

At a high level the RNNFast chip consists of Global Memory, a Computational Memory array, Configuration Memory and I/O interface as shown in Figure \ref{Fig.:architecture}. The Global Memory is a dense memory block implemented using DWM. This is the main memory of the accelerator and is used to store inputs and results. The Computational Memory is the compute engine and is implemented primarily using DWM elements augmented with CMOS logic where appropriate. The compute array is organized as a pool of highly reconfigurable and tightly interconnected tile groups. 

One or more multi-layer RNN networks can be mapped to multiple tile groups, in a weight-stationary design (weights are stored locally in the Computational Memory). The Configuration Memory holds the runtime configuration settings for the chip. RNNFast is optimized to deliver low latency without batching, and it is also efficient for batch workloads. 

\subsection{Compute Tiles}
\begin{figure}[ht]
\centering
\includegraphics[width=0.7\linewidth,trim = 0 20 0 30]{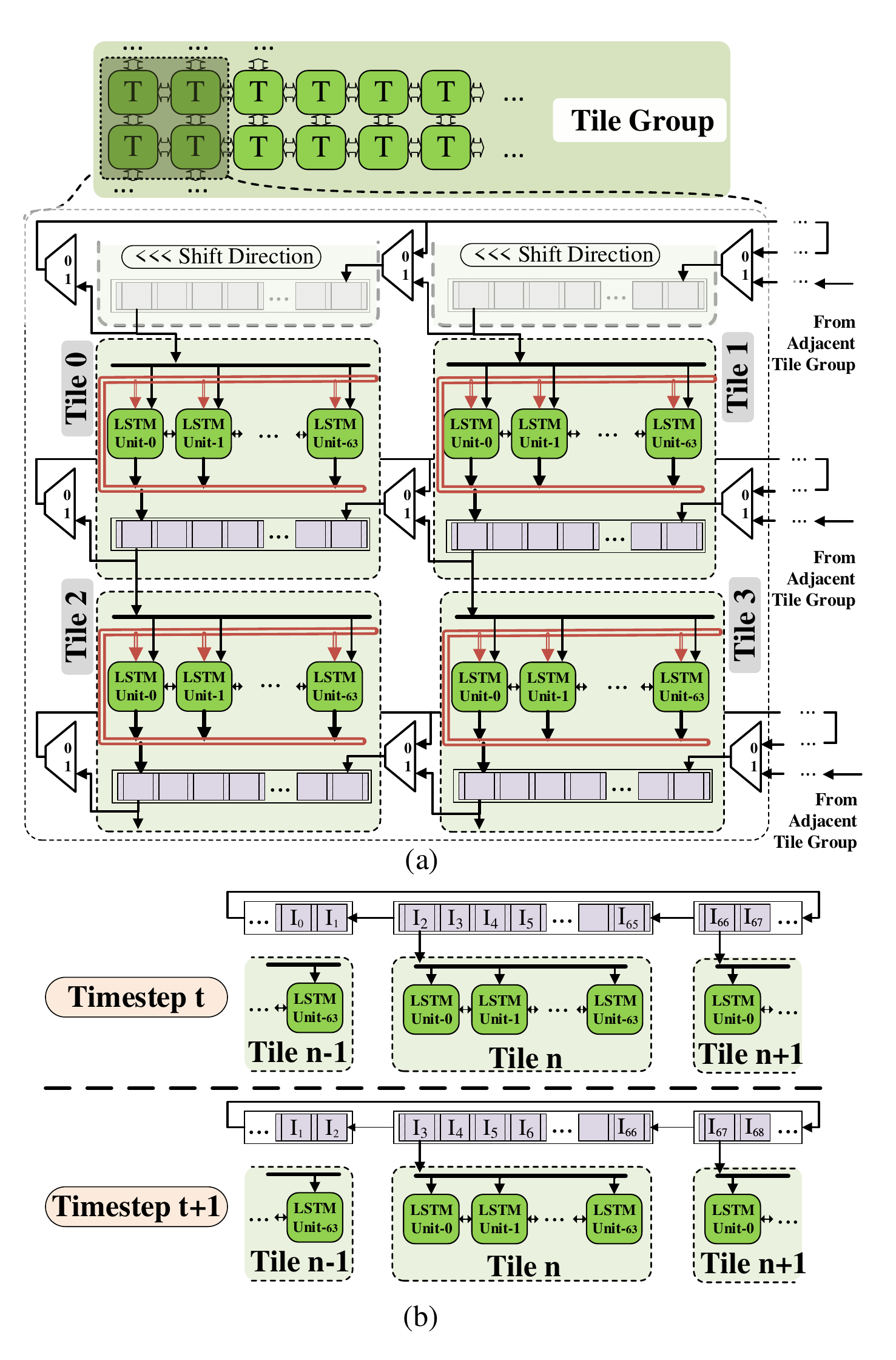}
\caption{\small (a) Compute tile layout, internal design and interconnection through racetrack chains. (b) Reading inputs into tiles in two consecutive timesteps. 
}\label{Fig.:tile}
\vspace{-1em}
\end{figure}

A compute tile consists of multiple LSTM hardware units that share a single input and a single output racetrack. They are interconnected with their nearest horizontal and vertical neighbors through racetrack memories. Figure \ref{Fig.:tile} shows the tile design and layout. The results of the computation within each tile are written directly onto the input track of the tile belonging to the next layer in the network. Tiles are organized in tile groups, which are connected to each other through traditional wired interconnection networks. 

\subsubsection{Inter-tile Communication}

RNNs are typically fully connected networks requiring  all inputs to be delivered to all the neurons in a given layer. The high degree of connectivity that has to be supported by the hardware can lead to substantial energy and area overheads when traditional wired interconnects are used. To address this challenge we leverage the shifting mechanism of DWM racetracks for communication both within and across tiles. 

Within a tile, inputs are read sequentially from the tile's input racetrack and broadcast to all LSTM units across a locally-shared bus. Each read is followed by a shift of the input track to align the next input element with the read head. Figure \ref{Fig.:tile} (b) illustrates two timesteps in this process. In addition to the tile-local broadcast, each input is also sent to the neighboring tile on the left for addition to its input track. We call this process "chaining". Chains are essentially circular buffers that circulate all inputs to all tiles that are mapped to the same layer of the NN. Chains of different lengths can be configured depending on the number of neurons in each layer of the network. Racetracks are connected through MUXs (Figure \ref{Fig.:tile} (a)) that enable different chain lengths. A variable number of tracks can be included in a chain by simply setting the right most track MUX to 0 and the rest to 1. 
 
\subsection{LSTM Units}
\label{sub:LSTM}
\begin{figure}[t]
\centering
\includegraphics[width=0.7\linewidth,trim = 0 0 0 0]{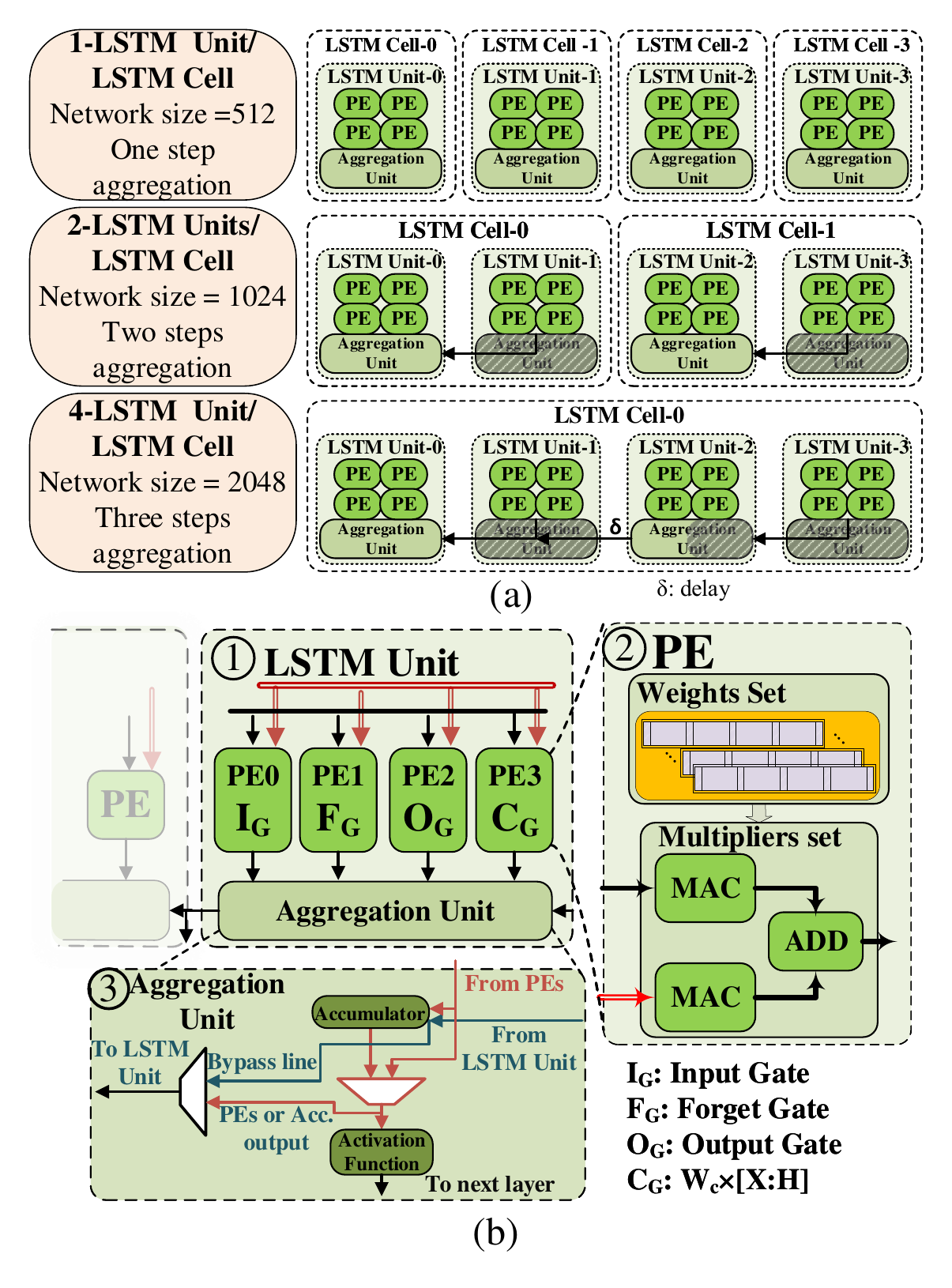}
\caption{(a) Three mapping examples of logical LSTM cells to LSTM units. (b) LSTM unit design.  }
\label{Fig.:sublstm}
\vspace{-1em}
\end{figure}

Each tile consists of multiple LSTM compute units (64 in our design). RNNFast is a weight-stationary design, with fixed capacity for weight storage in each LSTM unit. A logical neuron can be mapped to one or more LSTM compute units depending on the number of weights it requires. We expect a 1-to-1 mapping between logical neurons and hardware LSTM units for most networks. However, when a logical neuron requires more weights than a single LSTM units can store, it is mapped to multiple LSTM units. Figure \ref{Fig.:sublstm} (a) shows three mapping examples for a single logical LSTM cell: 1 LSTM unit (top), 2 LSTM units (middle) and 4 LSTM units (bottom).


\subsubsection{Processing Elements}
\label{subsub:PE}
The architecture of an LSTM cell is shown in Figure \ref{Fig.:sublstm} (b). Each cell is subdivided into four processing elements (PEs) \circled{\color{black}1}. Per equations \eqref{it_eq} -- \eqref{ht_eq}, each input $X_t$ is multiplied with four different sets of weights. A single PE can be assigned to any one of the  weight sets (known as gates), e.g. $I_{G}$, $F_{G}$, $O_{G}$ or $C_{G}$. However, an LSTM cell gate can be mapped to one or more PEs across LSTM units depending on its storage requirements and input/output fanout. Allocating four hardware PEs to each LSTM unit allows RNNFast to accommodate different RNN variants (see Section \ref{subsec:mapping}). 

PEs have racetrack-based storage for weights and racetrack-based compute units, including multiply accumulator (MAC) engines for matrix multiplication. The MAC engine is composed of 256+16 DWM based full adders. The MAC unit is deeply pipelined into 48 stages. In order to increase parallelism, each PE uses two MAC engines, one for the main input $X_t$ and one for the feedback input $h_{t-1}$. 

Each PE unit holds a set of weights and performs the dot product on the corresponding subset of inputs. Each PE only consumes inputs corresponding to the weights it stores. Each input to a PE is multiplied by its weight and accumulated with the result of the previous multiplication \circled{\color{black}2}. Each PE stores the result of the accumulation in its own output racetrack.

\subsubsection{Input and Weight Mapping}
\label{sec:sec:layout}
\begin{figure}
\centering
\includegraphics[width=0.7\linewidth,trim = 0 0 0 0 0]{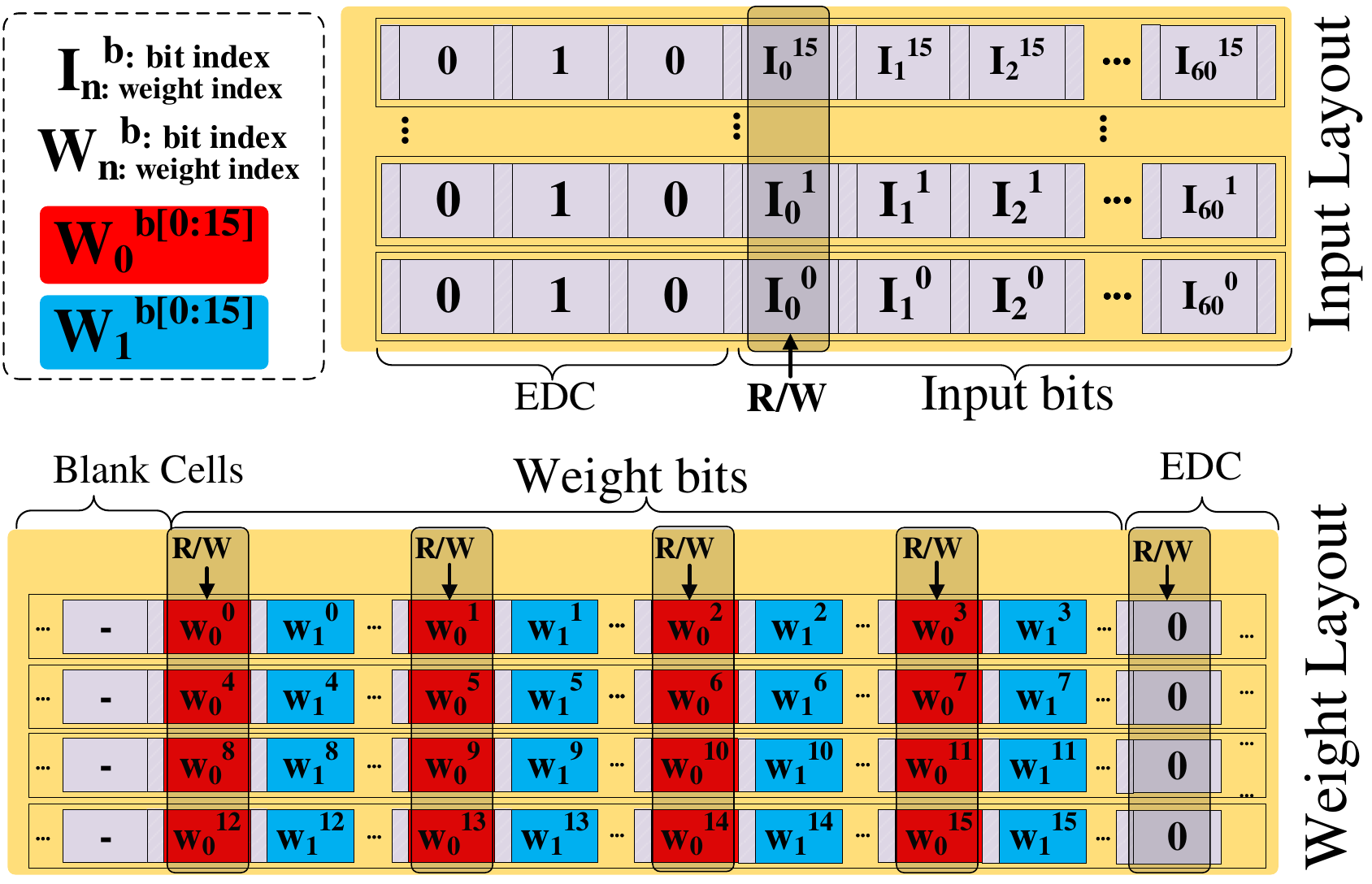}
\caption{Mapping of inputs and weights to racetracks.}
\label{Fig.:layout}
\vspace{-1em}
\end{figure}

The inputs and weights assignment to racetracks is a trade-off between access latency and hardware overhead. In RNNFast, inputs are spread across multiple racetracks with 1 bit per track. This allows an entire input word to be read in a single cycle, as the top half of Figure \ref{Fig.:layout} illustrates. Error detection bits are also included in the tracks and their role will be detailed in Section \ref{sec:errors}. Note that the input tracks do not require dummy domains (Figure \ref{Fig.:tile}-b). Values at the end of the track are read and sent to the neighboring track.

Unlike inputs, which move from track to track along the chain, weights are stationary at PE level and are reused multiple times. This means that after scanning all weights, the tracks need to be returned to the initial weight. To minimize the number of shifts, weight values are distributed both within and across multiple racetracks. Weight racetracks are provisioned with multiple read/write heads (5 in our design) which divide the racetrack into 6 10-bit segments. The left-most segment domains are used as dummy domains and the rest of the segments are used to store weight values. Data layout is such that all read heads across all tracks can access all the bits of a single weight simultaneously. Racetracks are grouped in sets of 4, with each set storing 10 weights. The bottom of Figure \ref{Fig.:layout} illustrates this layout. Weight $W_0$ (red) is currently aligned with the read heads. A single-position shift to the left will align the next weight $W_1$ (blue) with all the read heads. Access to each set of weight racetracks is pipelined. When all 10 weights are read from the current set of racetracks, the next set of weights will be read from next set. While the new weights are accessed, the weights in previous set are shifted back to their initials positions. This takes place when the racetrack set is not being accessed and is therefore off the timing critical path.


\subsubsection{Result Aggregation}
\label{subsub:Aggre}
If more than one LSTM unit is mapped to a neuron the partial results of the individual LSTMs have to be combined to form the neuron's output. Aggregation units \circled{\color{black}3} in each LSTM are used to sum up partial results in that LSTM block. In addition, the aggregation units apply the sigmoid and tanh functions and perform the multiplication and accumulation operations in order to generate the final output of the cell.

For cases in which neurons span multiple LSTM blocks, aggregation units in those blocks are linked to produce the final result. This is achieved by collecting all the partial results computed by each LSTM unit (mapped to the same neuron) to a single aggregation unit. Aggregation units are also chained through adjacent LSTM units. Each aggregation unit sends out its final result to the adjacent aggregation unit to its left. The adjacent unit will use the incoming result to either accumulate or bypass it to the next unit (Figure \ref{Fig.:sublstm}-\circled{\color{black}3}). Even-indexed aggregation units consume and odd-indexed aggregation units forward the incoming result. The leftmost LSTM in a neuron will be responsible for the final aggregation and will apply the sigmoid and tanh. Aggregation time is a logarithmic function in the number of LSTM cells mapped to a single neuron. This is also done by setting multiplexers in the aggregation unit and power gating the inactive units in output generators at odd indexed LSTM units.

The design tradeoff for LSTM units is driven by the need to support networks both large and small. If LSTM units and PEs are too large, storage space will be wasted when small networks are mapped. If they are too small, large networks will require several LSTM units per neuron, increasing the aggregation time.

\subsection{Nonlinear Functions}
The nonlinear functions are an important component of the RNN cells and are used for output activation.
RNNFast uses hardware acceleration for the sigmoid and tanh nonlinear functions. The hardware is included in each Aggregation Unit (Figure \ref{Fig.:sublstm}). We propose an area efficient approximate logic function-based unit implemented using DWM for the nonlinear functions. 

\begin{figure}
\centering
\includegraphics[width=0.6\linewidth,trim = 0 0 0 0 0]{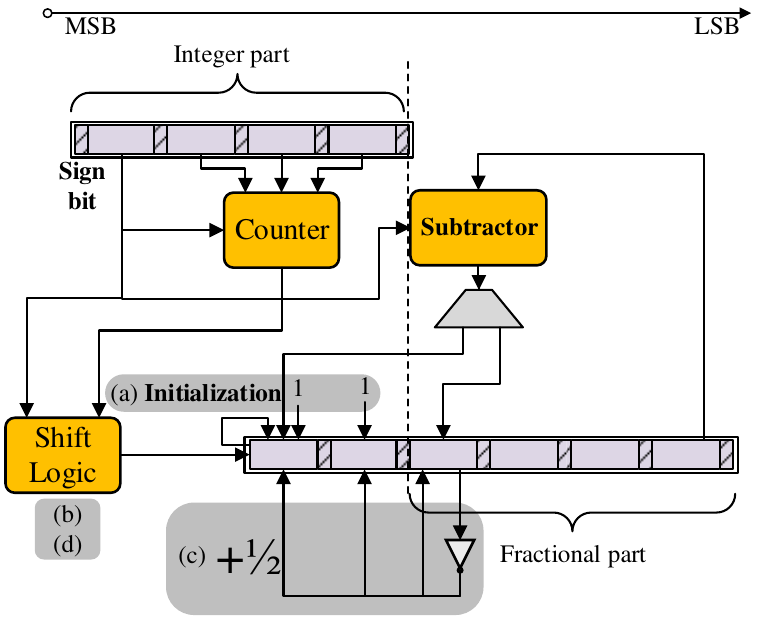}
\caption{DW based implementation of sigmoid/tanh.}\label{Fig.:approx}
\vspace{-1em}
\end{figure}
The approximation has been proposed by prior work \cite{tommiska_cdt2003} as an alternative to the standard sigmoid follows Equation \ref{approx_sig_eq}: 
\begin{equation} \label{approx_sig_eq}
\sigma\left(z\right) = 
	\begin{cases}
		\frac{\frac{1}{2}+\frac{\hat{z}}{4}}{2^{\mid (z)\mid}}~~~~if z < 0\\
		\\
		1-\sigma \left(-z\right) ~~if z > 0 \\
	\end{cases}
\end{equation}
This approximation has the advantage of being easier to implement in hardware. As Equation \ref{approx_sig_eq} shows, the hardware has to support division by $2^n$ numbers. This can be implemented using shift operations which are a feature of racetrack memories. The tanh approximation function can be computed from the sigmoid function through two multiplications and a subtraction.
Note that $\hat{z} = z + \mid(z)\mid$, where $(z)$ is the integer part of $z$.

Figure \ref{Fig.:approx} shows our DWM-based implementation of the sigmoid approximation. Sigmoid for a negative value will be computed as follows: a) the output integer part is initialized with binary '1'; b) two right shifts are performed to compute $\hat{z}$/$4$; c) $+1$/$2$ is applied to the result; d) final result is shifted right $\mid (z)\mid$ times.
For a positive number two subtraction steps are added in the beginning and end of above steps. To compute the tanh approximation, a right shift ($2\times{z}$) and a subtraction will be applied in the first and last steps respectively. 
This design is very area and energy efficient utilizing only a 16 bit racetrack memory, along with some simple subtraction and counting logic. Section \ref{sec:evaluation} evaluates the relative merits of the approximate designs regarding LUTs. 

\subsection{RNNFast Mapping and Configuration}
\label{subsec:mapping}
The RNNFast hardware can be configured to implement different network sizes and topologies. Moreover, multiple distinct neural networks can be mapped to the same chip.

\begin{figure}
\centering
\includegraphics[width=0.6\linewidth,trim = 0 0 0 15]{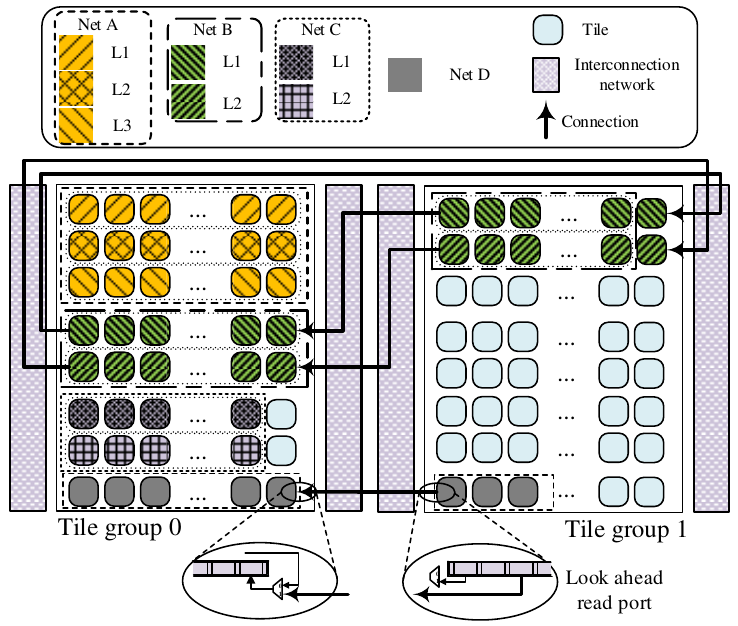}
\caption{\small 
Mapping multiple LSTM networks to RNNFast. Interconnection network helps extend racetrack chains beyond tile groups for large networks.
}\label{Fig.:net-map}
\vspace{-1em}
\end{figure}

Outputs from one network can be delivered directly to the following network or stored in the on-chip memory for further processing, if needed. Figure \ref{Fig.:net-map} illustrates an example of four networks A, B, C and D mapped to two tile groups. Tile groups are connected through a wired interconnect. The racetrack chains for each row of tiles have additional read/write heads to provide access to the inter-tile network.

Multilayer networks span multiple rows with different layers mapped to consecutive rows. Tile groups are designed with wide rows to accommodate most network sizes (e.g. Nets A and C). However, when a network layer cannot fit in a single row, RNNFast supports splitting it across tile groups (e.g. Nets B and D). This is achieved by extending the input/output racetrack chains to neighboring tile groups using the inter-group wire interconnect. We chose to split layers across tile groups (as opposed to within a tile group) in order to allow consecutive network layers to continue to be mapped to adjacent rows, preserving inter-layer communication.  

One important design constraint was to enable the extension of the racetrack chains across tile groups without adding to the track chain shift latency. This is accomplished by implementing a look-ahead read port at the end of the track that reads inputs several cycles ahead of the end of the track, as illustrated for Net D in Figure \ref{Fig.:net-map}. This allows the input to reach the destination row in the neighboring tile through the higher latency interconnect by the time the same input reaches the end of the source track.

\subsubsection{Other LSTM Variants}

\begin{figure}
\centering
\includegraphics[width=0.6\linewidth,trim = 0 0 0 20]{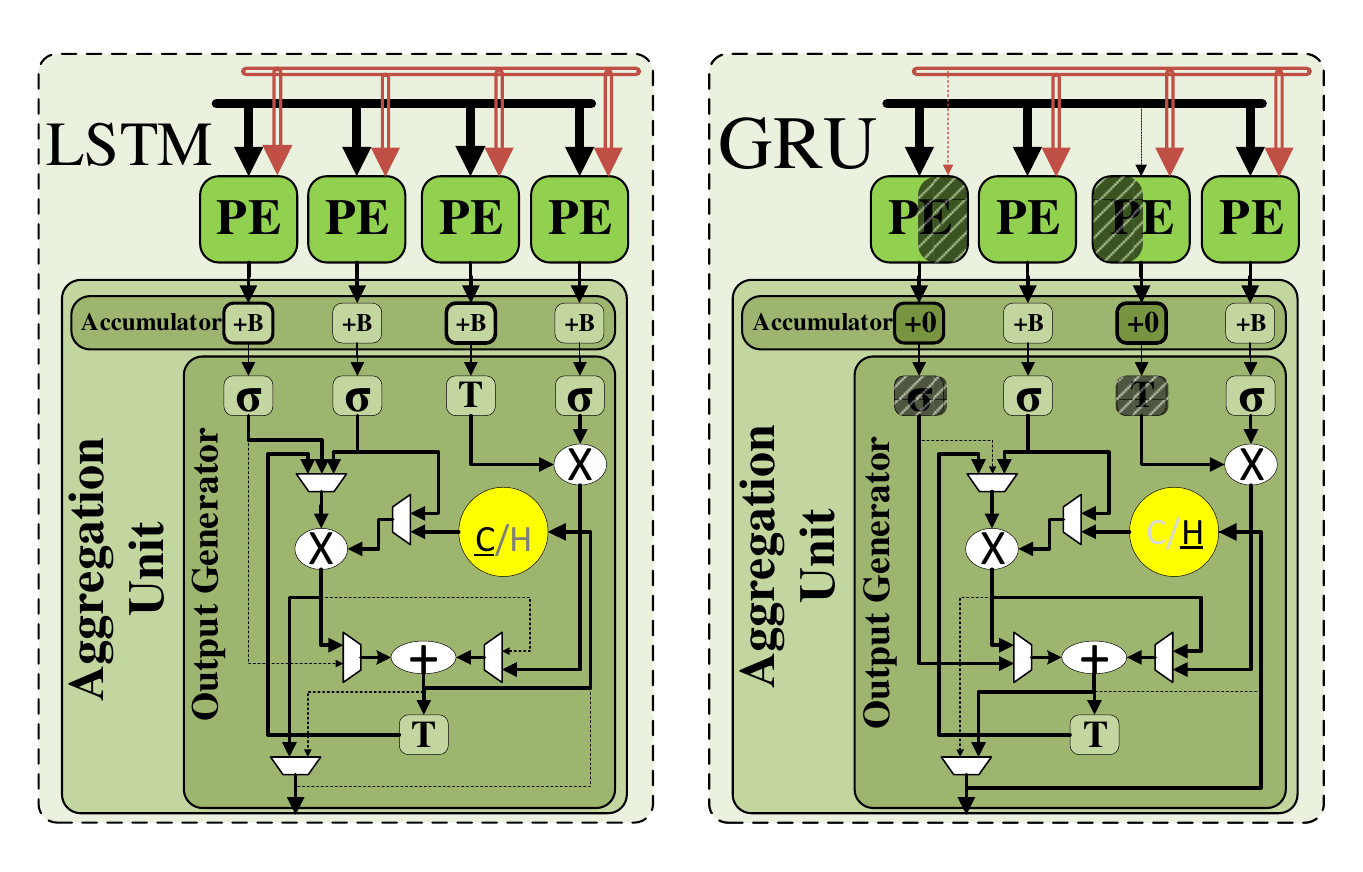}
\caption{LSTM vs GRU cell configuration on RNNFast}\label{Fig.:GRU}
\vspace{-1em}
\end{figure}

RNNFast is designed for the more demanding LSTM design. However it is also compatible with LSTM variants like Gated Recurrent Unit (GRU) and Vanilla RNN, which require fewer compute resources. Unlike LSTM, the GRU unit does not use a memory element to control the flow of information and are useful when input sequences are not very long. Figure \ref{Fig.:GRU} shows how a GRU cell can be mapped to a RNNFast LSTM unit. The shaded areas represent unutilized components. GRU utilizes  75\% of the MAC resources. 


Simpler RNNs like \textit{Vanilla RNN}, only utilize a single PE per neuron and do not need the aggregation unit. As a result, RNNFast can map four \textit{Vanilla RNN} neurons in each LSTM unit.

Moreover, RNNFast allows the mapping of other network types such as  Bidirectional RNNs (BiRNN). A BiRNN consists essentially of two RNNs stacked on top of each other. The output is computed based on the hidden state of both networks. In our design the two networks are mapped on the hardware in an interleaved fashion. The aggregation hardware is used to link the two networks. The input data is also duplicated and interleaved in reverse order ($x_1,x_n,x_2,x_{n-1},x_3,x_{n-2},...,x_n,x_1$).

\subsubsection{RNNFast Configuration}

The RNNFast configuration is programmed through configuration registers that control input assignment at PE level, input track chaining, result aggregation setup, etc. A configuration file with the LSTM network(s) specifications is loaded into the device driver of the accelerator and propagated to the appropriate registers.

\section{Error Mitigation Design}
\label{sec:errors}

\subsection{DWM Position Errors}
\label{sec:pos_errors}
As detailed in Section \ref{sec:background_dwm},  "out-of-step" shift errors (in which the wrong bit is aligned with the read/write heads) are a significant reliability challenge for DWM. Since RNNFast accesses data sequentially, that means virtually all accesses require only single-position bit shifts, we focus on single-bit overshift errors which are expected to occur with a probability of $10^{-5}$ \cite{Zhang_ISCA_15}, which is quite high. We used Pytorch \cite{pytorch} to inject error in weights for both im2txt and seq2seq models.

\begin{figure*}[!htb]
\minipage{0.49\textwidth}
 \includegraphics[width=\linewidth]{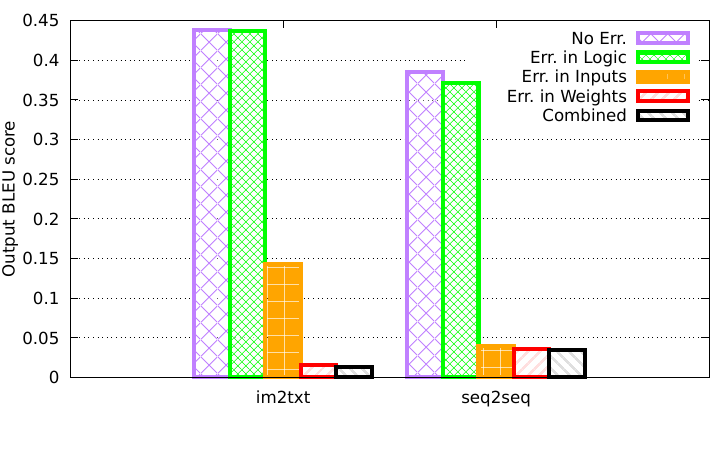}
\caption{\small Output accuracy (BLEU score) for logic, inputs and weights components.}\label{Fig.:error_fig1}
\endminipage\hfill
\minipage{0.49\textwidth}
 \includegraphics[width=\linewidth]{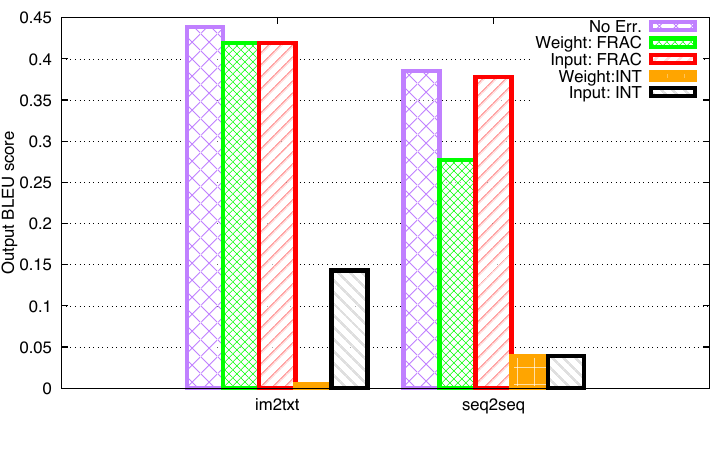}
\caption{\small Output accuracy (BLEU score) for integer and fraction components.}\label{Fig.:error_location}
\endminipage
\end{figure*}

While prior work \cite{reagenisca2016} has shown that neural networks are quite resilient to errors, we find that error rates on the order of DWM overshift errors can degrade output accuracy substantially. Figure \ref{Fig.:error_fig1} shows the accuracy of the output for two benchmarks, measured by the BLEU (bilingual evaluation understudy) metric \cite{Papineni_BMA02}, relative to an error-free baseline. 
BLEU is an algorithm for evaluating the quality of text which has been machine-translated from one natural language to another. Quality is considered to be the correspondence between a machine's output and that of a human. The models that we used have reported very close BLEU scores to the state of the art models \cite{toral2018level}.
We inject single-bit overshift errors in different DWM components of RNNFast: the racetrack chains used to hold inputs and outputs for each NN layer, the weights associated with all PEs, the DWM components of the logic functions (MAC units and the nonlinear functions). Shift errors are modeled as a uniform distribution with an overshift probability of $4.55\times10^{-5}$ \cite{Zhang_ISCA_15}. 

Figure \ref{Fig.:error_fig1} shows that when errors are injected only in the logic, the drop in output accuracy is very low: <1\% for {\em im2txt} and 3\% for {\em seq2seq}, two of the benchmarks we run. This is because overshift off-by-one errors in the MAC and nonlinear functions tend to produce results that are relatively close to the correct value. As a result, the accuracy of the output is very high. However, when errors are injected into the input chains and the weight arrays, the output accuracy drops dramatically to between 10\% and 35\% of the original. When errors are injected uniformly in all DWM tracks, the output accuracy drops below 5\% for {\em im2txt} and below 10\% for {\em seq2seq}, meaning that the results are essentially useless. This data highlights that mitigation solutions for errors in the inputs as well as weights are essential. 

To better understand which errors have the worst effect on output quality, we selectively inject errors into different bits of data words. RNNFast uses 2's complement fixed point representation for both inputs and weights. We inject errors separately into the integer and the fraction portions of the word. Figure \ref{Fig.:error_location} shows the results of this experiment. When errors are injected only in the fraction, the drop in accuracy is less than 3\% for both inputs and weights in {\em im2txt}. For {\em seq2seq} the accuracy degradation is worse when errors are injected in the weights compared to inputs, but overall output quality is still reasonably high. 

Injecting errors with the same probability in the integer portion of the data words has a much more dramatic effect, leading to a drop in output accuracy of between 35\% and 10\%. The large effect is due to the fact that in these workloads both inputs and weights are represented with small fractional numbers. A single bit flip of the integer fraction can turn a small number into a much larger value, which has a disproportionate effect on the rest of the network.

The large effect on output accuracy is caused by the fact that due to 2's complement representation, a single shift error in a data word storing a small value can cause that value to be interpreted as a large value with the opposite sign. For example the binary "00000011.10000010" (3.5078125 in decimal) would flip into "00100011.10000010" (35.5078125) or "10000011.10000010" (-124.492188) when a non-sign or sign bit in integer part inverted, respectively. This is also true for a negative number, "11111111.00101010" (-0.8359375) turns into  "01111111.00101010" (127.1640625) after a sign bit flip.

\subsection{RNNFast Error Mitigation}
\label{sub:Error_mit}

RNNFast addresses overshit errors by implementing an efficient error mitigation mechanism that considers the sensitivity of RNN workloads to errors that result in very large values. We implement different error detection and mitigation mechanisms for input/output racetrack chains and for weight arrays. We take advantage of their design characteristics to implement a more efficient SEDSEC design that has lower area overhead, requires fewer extra domains and access ports compared to prior DWM EDC solutions such as \cite{Zhang_ISCA_15}.



\subsubsection{Input Errors}
In order to detect overshit errors in the input tracks, we append a 3-bit pattern to the left side of each track, as shown in the example in Figure \ref{Fig.:inputECC}. The figure shows a single track that stores bit {\em n} for multiple inputs $I_{1}-I_{7}$. In the initial state the Error Detection Code (EDC) "101" is stored in the leftmost bits of the track. Input $I_{1}$ is read in the current cycle. At time $t_{1}$ the track is shifted left by 1 to access the next input. If the shift is correct, the leading (check) bit should be a "1". Input $I_{2}$ is read and sent to the LSTM units. A new EDC code is written at cycle $t_{3}$ in the first three bits of the track using three parallel write ports. Note that updating the EDC does not introduce any time overhead since a write cycle already exists following each read to allow data to be written into the next track in the chain.

At cycle $t_{4}$ we show an overshift error. The track has incorrectly shifted left 2 positions instead of 1.  This means that $I_{3}$ (instead of $I_{2}$) is now aligned with the read head. The check bit is now "0" indicating a shift error. To recover from this error we use an additional read head to also read $I_{2}$. The outputs of the two read heads are connected to a multiplexer. The check bit value selects the multiplexer output (shown in blue in Figure  \ref{Fig.:inputECC}). A "1" selects the error-free output and a "0" selects the overshifted output. A similar mechanism selects the correct location for writing the input coming from the previous track in the chain. If an overshift error occurs, the write location is also shifted to the left, as the right hand side of Figure \ref{Fig.:inputECC} shows. 

At $t_6$ the EDC code is again updated. Following an overshift error the shift controller will not issue a shift command for the following cycle ($t_7$) since the track is already properly aligned to access the next input ($I_4$) during that cycle. Note that, since individual words are stored across multiple tracks to enable single-cycle access, an overshift error will affect all inputs that share that track (up to 60 in our design). It is therefore important to detect and correct these errors.

\subsubsection{Errors in Weight Arrays}
A similar mechanism is deployed to detect and mitigate errors in weight arrays associated with each PE. However, because the access timing to the weights array is more critical and weights are stored in a more compact representation, the detection and mitigation steps are implemented differently. Unlike inputs, which move from track to track along the chain, weights are stationary at PE level and are reused multiple times. This means that after scanning all weights, the tracks need to be returned to the initial weight. To minimize the number of shifts, weight values are distributed both within and across multiple racetracks. Weight racetracks are provisioned with multiple read/write heads (5 in our design). Data layout is such that all read heads across all tracks can access all the bits of a single weight simultaneously.

\begin{figure}[!htb]
\centering
\minipage{\linewidth}
\centering
 \includegraphics[width=0.6\linewidth,trim = 0 0 0 0 0]{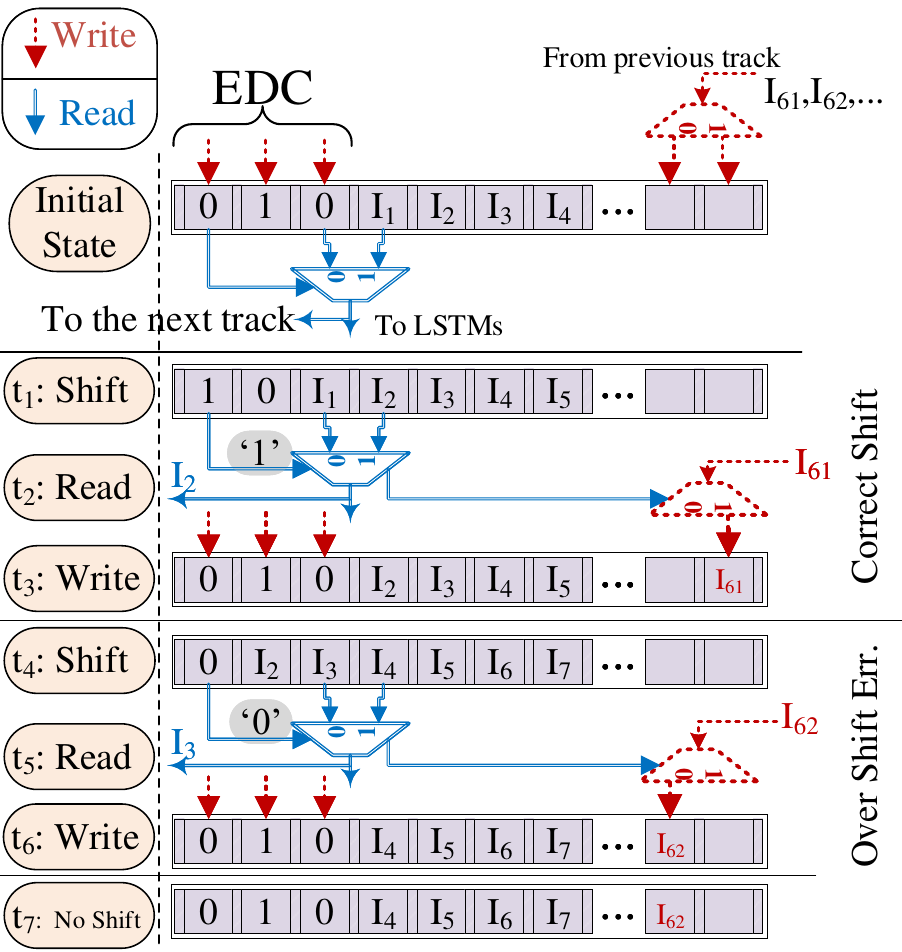}
\caption{\small Mitigation mechanism for overshift errors in the input track chains.}\label{Fig.:inputECC}
\endminipage\hfill
\minipage{\linewidth}
\centering
 \includegraphics[width=0.75\linewidth,trim = 20 0 0 0 ]{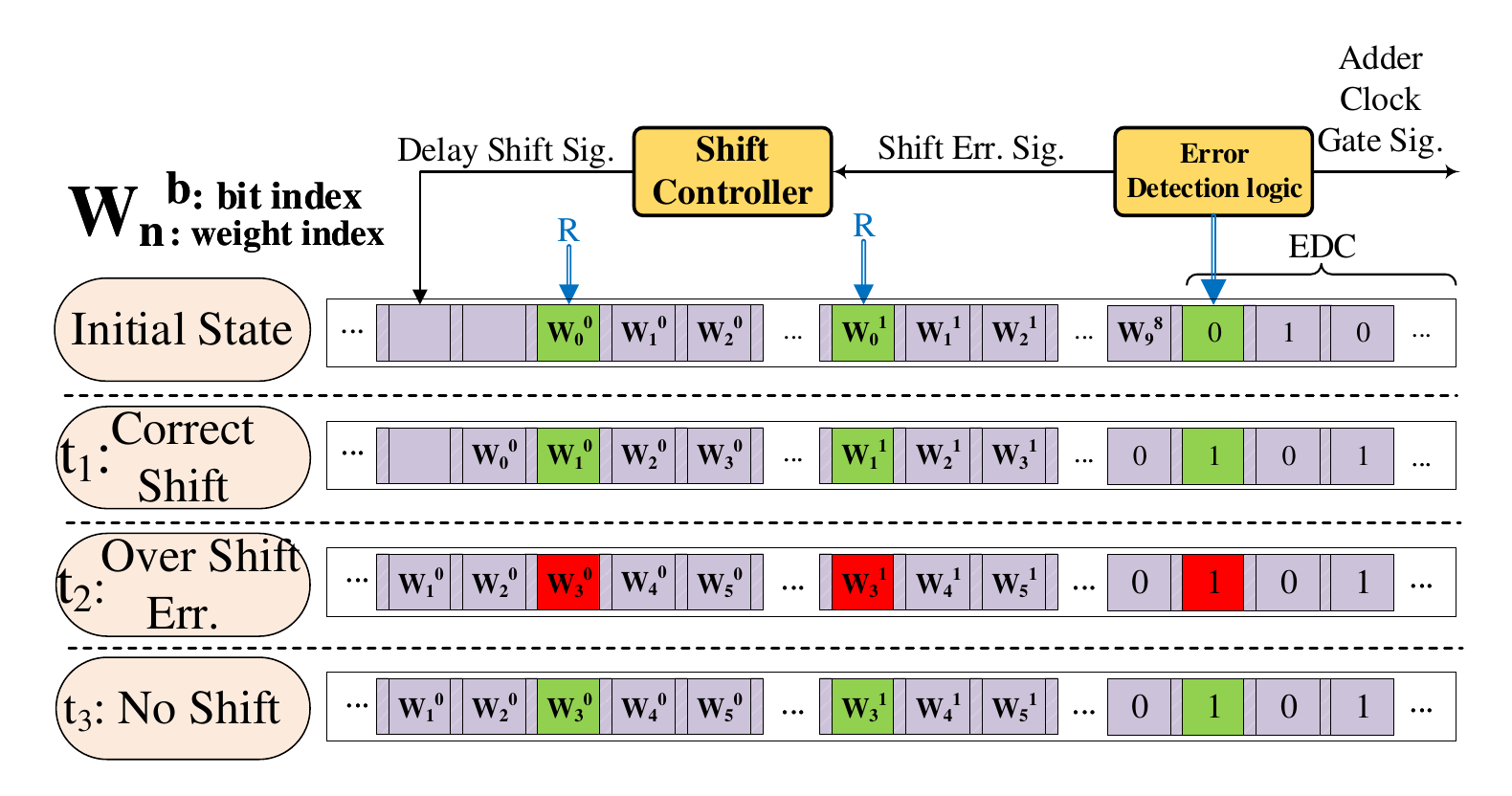}
\caption{\small Mitigation mechanism for overshift errors in the weight track chains.}\label{Fig.:error_det}
\endminipage
\vspace{-1em}
\end{figure}

Also, unlike the input racetrack chain, access to the weight arrays does not require a write cycle, so an update to EDC code is not feasible. We instead store a fixed EDC pattern of "01010" at the rightmost edge of the weight tracks as shown in Figure \ref{Fig.:error_det}. Error detection logic detects an overshift error when the current EDC bit does not match the expected value. For instance, in the initial state, the read heads are aligned with bits from weight $W_0$ and the error detection logic expects to read "0" from the EDC. 

At time $t_1$ a correct shift takes place and $W_1$ can be read. At time $t_2$ an overshift error occurs and weight $W_3$ is read instead of $W_2$. A recovery mechanism similar to the one for inputs could be employed. This would require doubling the number of read heads in each track and extra logic. Since weight storage in RNNFast is substantial, the overhead would be nontrivial. We can, however, avoid this extra overhead by leveraging the observation that replacing the incorrect weight with "zero" yields very little loss in output accuracy compared to error-free execution. This is in contrast with using the erroneous weight, which can be a large value. The following cycle at $t_3$, the shift controller will not shift because the track is already aligned for accessing the next weight.

\section{Evaluation Methodology}
\label{sec:method}

\subsection{RNNFast Modeling Infrastructure} 
\label{subsec:Infrastructure}

We implemented a detailed behavioral model to evaluate performance, chip area and energy consumption of the RNNFast design.  A cycle-level model that accounts for the latency of each component in the design is used for the timing simulation. The simulated hardware is configured for each neural network in our benchmark set, by enabling the appropriate number of hardware  tiles, LSTMs and PEs. Since all LSTM units execute independently and in parallel, only a single LSTM per tile is simulated to speed up simulation time. For the energy evaluation, the number of reads, writes, shifts as well as decoder, Adder/Multiplier and LUT accesses are counted for all the units in the design. 

To understand the energy consumption, shift and write latency of the Domain Wall Memory (DWM), an electrical model is necessary. A Verilog-A based SPICE model for DWM from \cite{motaman_vlsi2016,motaman_nanotech2015,motaman_islped2014} was simulated on Cadence Virtuoso. The DWM model estimates the effective resistance as a function of the length of the track and uses width and thickness of the strip to calculate current density and position shift. A Cadence component was created for the DWM model and a test-bench was setup to stimulate the device. A sensitivity analysis was conducted to study the effect of track length on shift latency and energy.

Table \ref{tab:rnn-info} shows the characteristics of the DWM we model and also lists the architectural parameters for RNNFast and power/area breakdown for different components. As weight values are in 16-bits precision, each four set of racetracks stores 10 weights. For storing 512 weights each PE needs 205 racetracks (Table \ref{tab:rnn-info}). We performed energy analysis on the number of LSTMs per tile and chose the number of LSTM per tile as 64. More details are in section \ref{sec:tuning}. The number of accumulator, multiplier, sigmoid and tanh units in the Aggregation unit (figures \ref{Fig.:LSTM} and \ref{Fig.:GRU}) is optimized for energy and performance. The smallest number of units that allows the LSTM to operate without stall cycles is chosen.

\begin{table}[h]
\small
\centering
\scalebox{0.68}{
\begin{tabular}{|c|c|c|c|c|}
\hline
\hline
\multicolumn{5}{|c|}{\large DWM properties}\\
\hline
\multicolumn{2}{|c|}{racetrack width/length/thickness}&  \multicolumn{1}{c||}{1F / 64F / 3nm} & domain length& 1F\\
\hline
\multicolumn{2}{|c|}{number of bits per track}&  \multicolumn{1}{c||}{64} & Effective cell size & 2.56$F^2$ \\
\hline
\multicolumn{2}{|c|}{read/shift/write latency  }&  \multicolumn{1}{c||}{1ns / 0.5ns / 0.5ns  } & Technology node & 32nm\\
\hline
\multicolumn{2}{|c|}{read/shift/write energy  }&  \multicolumn{1}{c||}{0.39pJ / 0.24pJ / 9.6fJ}&&\\
\hline
\hline
\multicolumn{5}{|c|}{\large Tile properties}\\
\hline
Component & Configuration & Specification & Power(mW) & area(${\mu}m^2$) \\
\Xhline{1pt}
Input buffer    &   1 track/tile    &   16 stripes/track    &   2.59    &   2.68\\
                &   with EDC        &   64 cell/stripe      &           &       \\
\hline
LSTM unit       &   64 per tile     &   4 PEs/LSTM          &  9.74     &   2046 \\
                &                   &   1 Aggre./LSTM       &           & \\
\Xhline{1pt}
\textbf{Total tile} &   &   \textbf{256 PEs}        &   \textbf{626}  &  \textbf{0.130$mm^2$}  \\
                    &   &   \textbf{64 Aggre. Unit} &                   &  \\
\hline
\hline
\multicolumn{5}{|c|}{\large PE properties}\\
\Xhline{1pt}
MAC             &  2/PE         &   272 Adder           &   \multirow{3}{*}{2.43}   &   \multirow{3}{*}{422}      \\
\cline{1-3}
Weight array    &  2 track/PE   &   205 stripes/track   &           &           \\
                &   with EDC           &   64 cell/stripe      &           &   \\
\hline
\hline
\multicolumn{5}{|c|}{\large Aggregation Unit properties}\\
\Xhline{1pt}
Accumulator &   4/LSTM      &   -                           & \multirow{4}{*}{0.004}    & \multirow{4}{*}{356} \\ 
\cline{1-3}
Multiplier  &   2/LSTM      &  -                            &     & \\
\cline{1-3}
sigmoid     &   3/LSTM      &Approx. nonlinear func. design &     &    \\
\cline{1-3}
tanh        &   2/LSTM      &Approx. nonlinear func. design &     &  \\
\hline
\hline
\multicolumn{5}{|c|}{\large On-chip DW Memory}\\
\Xhline{1pt}
\multicolumn{5}{|c|}{\textbf{Size: 128MB, 4R/W ports, Area: 6.2$mm^2$, Acc. Eng.: 0.89nJ, Acc. lat.: 1.69ns,  Leakage 24.3mW}}\\
\hline
\end{tabular}
}
\caption{Racetrack memory and RNNFast design  parameters with associated power and area overheads.}
\label{tab:rnn-info}
\vspace{-1em}
\end{table}

\subsubsection{RNNFast Design Variations}
\label{subsub:RNNFast_var}
We compare our design with two alternative RNNFast architectures  that use CMOS and Memristor technologies. We call them RNNFast-CMOS and ISAAC-RNN respectively. 
For RNNFast-CMOS, we used SRAM buffers for both LSTM inputs and weight storage within PEs. MAC units are also implemented with CMOS logic. We used SRAM-based LUTs for the nonlinear functions. Input SRAM buffers are also chained like racetrack memories in order to deliver all inputs to all LSTM units. 

ISAAC-RNN is an ISAAC \cite{shafiee_isca2016}-like  design for RNN that stores inputs in eDRAM and is entirely CMOS and memristor-based. ISAAC-RNN uses 128x128 2-bit memristor crossbars, similar to what was used in ISAAC, for the dot product engine. We kept the input buffer and aggregation unit designs same as RNNFast in order to only see the effect of memristor in the design and have a more fair comparison since eDRAM and CMOS logic has higher energy consumption than DWM. Each memristor dot product engine is capable of $128\times16$ multiplications in parallel (128 inputs by 16 weights). In an LSTM neuron each input is multiplied by 4 different weight sets. Thus, each memristor dot product engine can handle 4 neurons, making each crossbar in ISAAC-RNN computationally equivalent to 4 LSTMs in RNNFast. Thus there are 16 LSTM units per tile for ISAAC-RNN instead of 64 per tile in RNNFast. Inputs go bit by bit to the memristor crossbars. However, a chuck of 128 inputs needs to be supplied in a single cycle. We changed the input layout to maximize the performance of ISAAC-RNN, for a fair comparison.

\subsubsection{GPU Baseline}

We choose as a baseline system for evaluation a GPGPU optimized for machine learning: the NVIDIA Tesla P100 (Pascal architecture) with 16GB of  CoWoS-HBM2 memory. All our benchmarks use
the DNN-optimized cuDNN NVIDIA libraries version 7 \cite{cudNN}, which delivers roughly 6$\times$ performance improvement relative to a standard GPU implementation for LSTM on Torch \cite{cudNN5}.
We measure runtime of the forward passes through the LSTM layers using instrumentation in Deepbench. We measure power consumption using the NVIDIA SMI profiler. Since the SMI provides total board power, in order to isolate the active power of the GPU we subtract power measured at GPU idle. Since the board components are less energy proportional with activity compared to the GPU, they will account for most of the idle power. 

\subsubsection{PUMA} We also compared our design with PUMA \cite{ankit2019puma} a recently proposed DNN accelerator built with ReRAM. The authors of PUMA released a simulator and toolchain that we use to compile and run our benchmarks. We used the PUMA compiler to find the number of tiles required for each benchmark. We then set the simulator configuration file to inference mode and used the PUMA simulator to measure runtime and energy consumption. 



\subsection{Benchmarks}

We used LSTM-based RNN workloads from the Deepbench \cite{deepbench} open source benchmark suite for DNNs, released by Baidu. For our experiments we used:

\begin{table}[h]
\small
\centering
\scalebox{0.75}{
\begin{tabular}{|l|l|l|l|l|l|}

\hline
\multirow{2}{*}{Bench.} & \multirow{2}{*}{Platform} & \multirow{2}{*}{Precision} & Layers$\times$ & Time- & \multirow{2}{*}{Description} \\
& & & Neurons &step &\\
\hline
im2txt & DeepBench & 16 bit& 1$\times$512 & 11 & image caption \\
\hline
seq2seq & DeepBench & 16 bit & 3$\times$1024 & 15& language translation\\
\hline
&       &   & 1$\times$512   &   &\\
mach-tran & DeepBench & 16 bit & 1$\times$1024& 25 & Machine translation \\
&   &   &   1$\times$2048    & &\\
\hline
lang-mod & DeepBench & 16 bit & 1$\times$1536& 50 & language modeling \\
\hline
D-Speech  &   DeepBench   &   16 bit & 1$\times$2816  & 1500 & Deep Speech\\
\hline
\end{tabular}
}
\caption{Summary of the benchmarks evaluated.}
\label{tab:benchmark-info}
\vspace{-1em}
\end{table}

\emph{Image Caption Generator:} This benchmark is based on
the ``Show and Tell''
Model \cite{vinyals_corr2016}, which is an encoder-decoder type neural
network. The decoder is an LSTM RNN
that generates captions from a fixed-length vector input. 

\emph{Sequence-to-Sequence Model:} This benchmark is based on the RNN encoder-decoder
model by Cho et al.~\cite{cho_corr_corr2014}, which performs language 
translation. The encoder and decoder are 3-layer LSTM networks.
\emph{Machine Translation:} also based on the RNN encoder-decoder model by Cho et al.~\cite{cho_corr_corr2014}.

\emph{Language Modeling:} a probability distribution over sequences of words. It is used in speech recognition, sentiment analysis, information retrieval and other applications \cite{ponte_sigir1998}.

\emph{Deep Speech:} a Speech-To-Text engine that uses a model trained by machine learning techniques, based on Baidu's Deep Speech research \cite{hannun_corr2014}.

All benchmarks are run using 16-bit precision arithmetic on both RNNFast and the P100 GPU.

\section{Evaluation}
\label{sec:evaluation}

We evaluate the RNNFast performance and energy consumption compared to the NVIDIA GPU, the CMOS-based and the Memristor-based RNNFast design. We evaluate the reliability of the RNNFast error mitigation. We show an area utilization estimate for different benchmarks. We also include a high-level comparison to other RNN accelerators. 

\begin{figure*}[!htb]
\minipage{0.49\textwidth}
 \includegraphics[width=\linewidth]{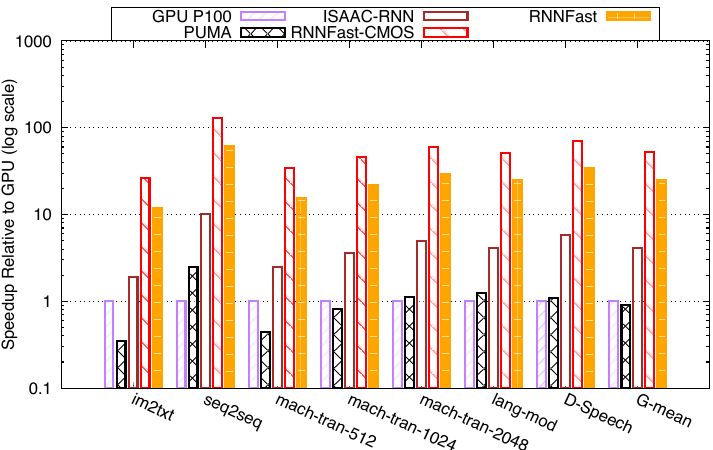}
\caption{\small RNNFast, RNNFast-CMOS, ISAAC-RNN and PUMA runtime relative to the GPU P100 execution.
\hspace{5cm}}
\label{fig:benchmark-performance}
\endminipage\hfill
\minipage{0.49\textwidth}
 \includegraphics[width=\linewidth]{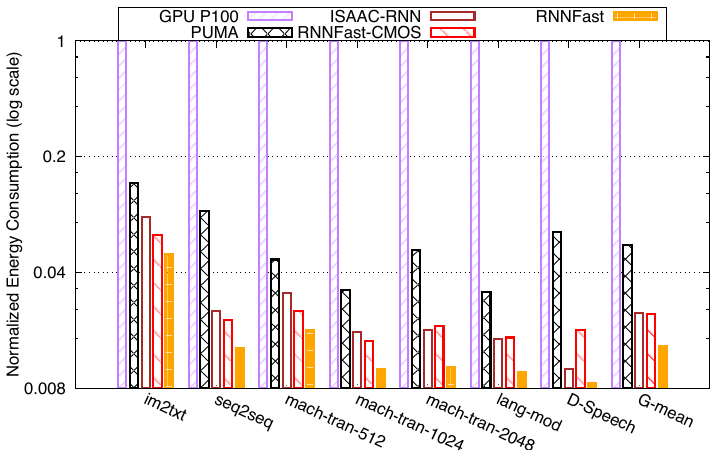}
\caption{\small Energy consumption for RNNFast, RNNFast-CMOS, ISAAC-RNN and PUMA relative to the GPU P100.}
\label{fig:benchmark-energy}
\endminipage\hfill

\vspace{-1em}
\end{figure*}
\subsection{Performance Improvement and Energy Saving}  
\label{sec:eval:performance}
Figure \ref{fig:benchmark-performance} shows the execution time speedup for RNNFast, RNNFast-CMOS and ISAAC-RNN relative to the P100 GPU for the seven benchmarks we run. RNNFast speedup relative to the GPU varies between 12$\times$ for {\em im2txt} and 34.5$\times$ for {\em D-speech}, with an average speedup of 21.8$\times$.
RNNFast speedups increase with the network size, demonstrating the excellent scalability of the design.
For instance, in {\em mach-trans} we test three different network sizes ranging from 512 to 2048, We observe speedups increases from 15.4$\times$ to 29.3$\times$. This is because the large number of threads required to handle the larger network becomes a bottleneck even for the GPU, whereas RNNFast scales much better.

ISAAC-RNN also brings a substantial speedup relative to the GPU ranging between 1.88$\times$ for {\em im2txt} and 5.8$\times$ for {\em D-speech}. Although this is significant, ISAAC-RNN is more than 6.1$\times$ slower than the DWM RNNFast implementation. This is primarily due to the higher latency of the LSTM unit in ISAAC-RNN, which is 7.3$\times$ higher than a RNNFast LSTM unit. The higher latency is due to the memristor array read latency (100ns) and overheads that stem from the ADC/DAC components. Even though a single memristor array can handle up to 4 neurons, which increases  throughput, ISAAC-RNN is still fundamentally slower than RNNFast.
RNNFast-CMOS shows $2.1\times$ speedup compared to RNNFast. This is due to faster CMOS adders and random memory access instead of the shift-based access in RNNFast.


The PUMA ReRAM-based design is more general that ISSAC and RNNFast, supporting both CNNs and DNNs. However, its performance is lower than both ISAAC-RNN and RNNFast. In general, PUMA tends to have better performance than the GPU for larger networks, especially for multi-layer networks (seq2seq) where PUMA benefits from its pipelined architecture.

Figure \ref{fig:benchmark-energy} shows the energy consumption for RNNFast, RNNFast-CMOS and ISAAC-RNN relative to the GPU in log scale. RNNFast reduces energy consumption on average by $70\times$. This is due to a much faster execution time achieved with about $1$/$3$ the power of a GPU. The RNNFast-CMOS design has $55\%$ higher energy compared to RNNFast. This is reaches a $100\%$ increase for {\em D-speech} due to higher resource demand, which increases the leakage energy for both compute and memory logic in CMOS. This causes the CMOS design to reach its maximum TDP for smaller networks. ISAAC-RNN also has higher energy usage than RNNFast due to its ADC/DAC and CMOS logic.PUMA energy consumption is much lower than the GPU, however, as expected is not lower than ISAAC-RNN. RNNFast is much more energy efficient, using about 25\% the energy of PUMA.

\begin{figure*}[!htb]
\centering
\minipage{0.49\textwidth}
 \includegraphics[width=\linewidth]{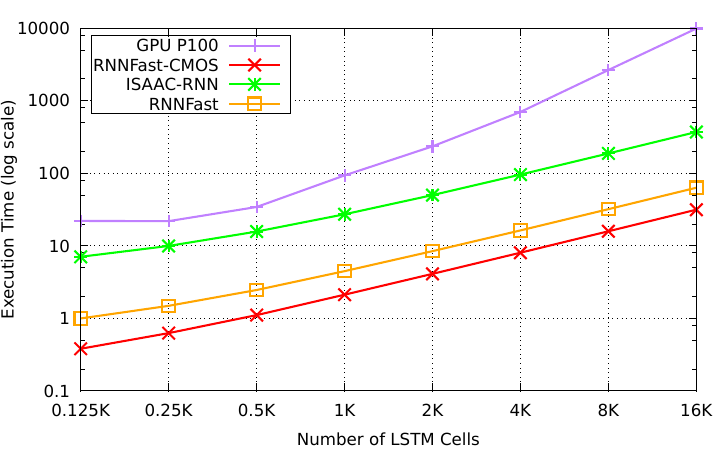}
\caption{ \small RNNFast, ISAAC-RNN and GPU execution times vs. net size for
  {\em mach-tran}, normalized to RNNFast 0.125K.}
\label{fig:deepbench-scalability}
\endminipage\hfill
\minipage{0.49\textwidth}
 \includegraphics[width=\linewidth]{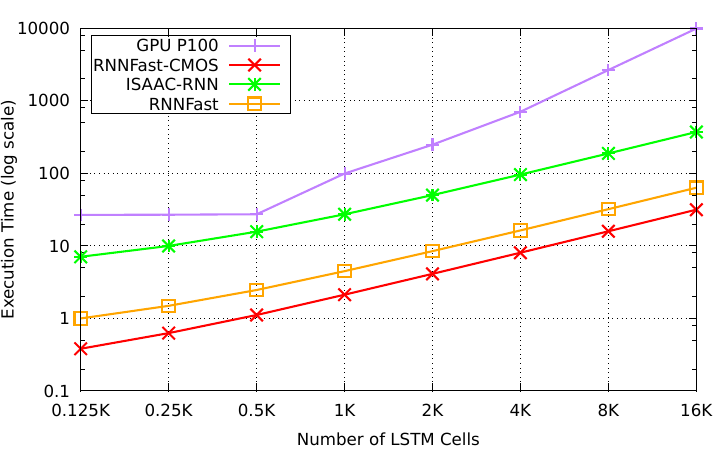}
\caption{  \small RNNFast, ISAAC-RNN and GPU execution times vs. network size for  {\em im2txt}, normalized to RNNFast 0.125K.}
\label{fig:seq2seq-scalability}
\endminipage\hfill
\vspace{-1em}
\end{figure*}

RNNFast offers a much more scalable design relative to a GPU due to its modular design and very high storage density of DWM. Figure \ref{fig:deepbench-scalability} shows the log scale of execution time for
the {\em mach-tran} benchmark as a function of problem (neural network) size ranging from 128 nodes to 16K nodes per layer in a single-layer configuration. For problem sizes larger then 16K, the GPU runs fail because the device runs out of memory. The GPU execution time exhibits a super-linear increase in execution time with problem size due to memory pressure. 
RNNFast is consistently faster than the GPU in the range of 13.9$\times$ (0.5K) to 156$\times$ (16K) and also scales better to very large problem sizes of 16K nodes and beyond. ISAAC-RNN scales similarly to RNNFast but it is also $6.2\times$ slower that RNNFast on average for {\em mach-tran}. RNNFast-CMOS shows almost $2\times$ speedup over RNNFast, at the cost of much higher energy.

Figure \ref{fig:seq2seq-scalability} shows a similar trend
for {\em im2txt}. The GPU shows good performance up to 0.5K, but run time increases exponentially beyond that. 

\subsection{Error Mitigation}
\label{sec:eval:errors}

We also evaluate RNNFast resilience to position errors. Figure \ref{Fig.:error_rate} shows the accuracy of the output as evaluated by the BLEU metric \cite{Papineni_BMA02}, as a function of the probability of position errors. We can see that for a relatively low probability of errors of $4.5\times10^{-7}$ the output accuracy is virtually unaffected. This is primarily due to the inherent robustness of the RNN to errors. However, without error mitigation, the output accuracy degrades substantially at higher errors rates. In the region around $4.5\times10^{-5}$ (highlighted region), which is the expected rate for single bit position errors, the output accuracy drops to 45\% for {\em im2txt} and 10\% for {\em seq2seq}, an unacceptable performance for most applications. When RNNFast error mitigation is enabled the drop in output accuracy is negligible at less than 2\%. 

The RNNFast error mitigation produces outputs with less than 5\% accuracy loss even for much higher error rates of $10^{-3}$ or around 20\% accuracy loss for  $10^{-2}$. This shows that RNNFast EDC is robust to much higher error rates than what is expected for DWM technology. 

It is also worth highlighting the fact that error mitigation incurs no performance penalty even when errors are detected. Correction or mitigation are performed without stalling the execution pipeline. This is an important design consideration because of the highly synchronized nature of the design. A single stall to correct an error would result in lost cycles for thousands of functional units.

\subsection{Nonlinear Function Hardware}

\begin{figure*}[!htb]
\centering
\minipage{0.49\textwidth}
 \includegraphics[width=\linewidth]{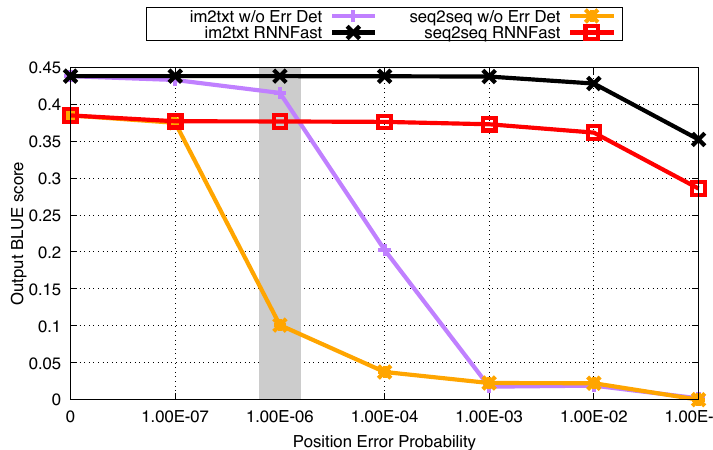}
\caption{\small Output accuracy for benchmarks} {\em im2txt} and {\em seq2seq} with and without RNNFast EDC.\label{Fig.:error_rate}
\endminipage\hfill
\minipage{0.49\textwidth}
 \includegraphics[width=\linewidth]{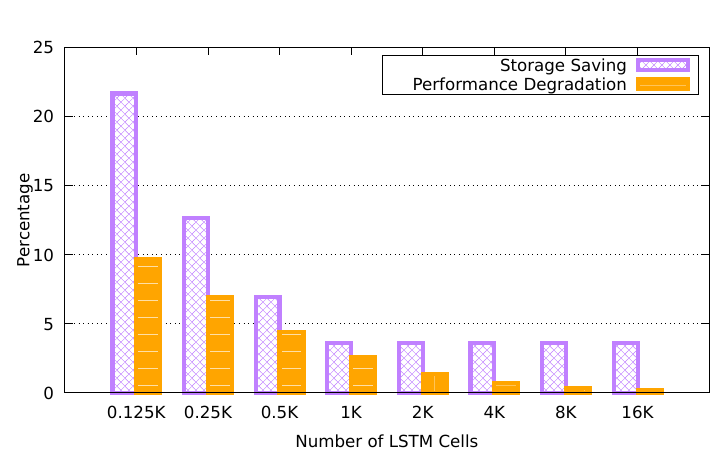}
\caption{ \small Storage saving and performance degradation for different network sizes for Approx. Function-based sigmoid design relative to LUT.}
\label{fig:storage}
\endminipage\hfill
\vspace{-1em}
\end{figure*}

We evaluate two designs for the nonlinear function hardware: a LUT-based implementation, and an approximate logic function-based unit. The function-based implementation is area efficient since it does not require as much storage as the LUT-based design. However the computation required, albeit simple, is slower than the simple lookup of the LUT version. The activation functions are not a significant latency bottleneck. However, at this scale we have thousands of such units on chip and reducing their area adds up to real savings. 
Figure \ref{fig:storage} shows the storage savings and performance degradation of the function-based sigmoid/tanh relative to the LUT design for multiple network sizes. The storage savings diminish as the network size increases because the storage space for the weights dominates. For large networks the storage savings are about 4\%, which represents >1GB of DWM for a 16K network. As for the performance cost, it starts at about 9\%, but falls below 1\% for larger networks. The approximated nonlinear function does not result in loss of accuracy as measured by the BLEU score.

\subsection{RNNFast Parameter Tuning}
\label{sec:tuning}
\begin{figure*}[!htb]
\centering
\includegraphics[width=0.5\linewidth]{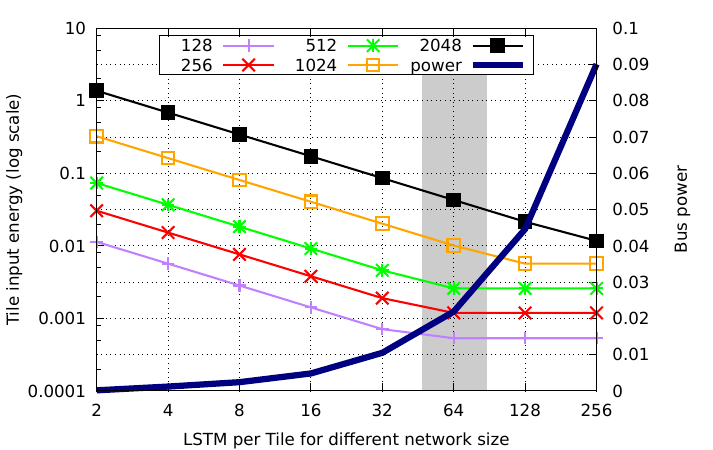}
\caption{\small Sensitivity analysis for the number of LSTMs per tile.}
\label{fig:LSTM_Tile}
\end{figure*}

We also conduct a sensitivity analysis on number of LSTM units per tile. Figure \ref{fig:LSTM_Tile} illustrates the tile input buffer energy versus different number of LSTMs per tile for different network size. As the number of LSTMs per tile increases, the power/area overhead for the within tile bus increases super-linearly. The minimum energy point is different depending on the size of the network. The 64 LSTM units per tile represents a reasonable compromise for medium-to-large networks. 

\subsection{Comparison to Other RNN Accelerators} 
\label{compare}

Several recent papers have proposed FPGA-based accelerators for RNNs \cite{sun2018fpga,li2018rnn,li2018fpga,wang2017accelerating,zhang2017power,Ferreira_16ReConfig,brainwave_isca18,mealey2018accelerating,han2017ese}. We provide a qualitative comparison with some of the more recent ones, for which runtime and energy numbers were available and similar applications were evaluated. Table \ref{tab:FPGA} summarizes the energy and runtime for FPGA-based designs from \cite{Ferreira_16ReConfig,mealey2018accelerating,han2017ese,brainwave_isca18} as well as the energy and runtime of RNNFast while running networks of equivalent size. 

The networks used in \cite{Ferreira_16ReConfig,mealey2018accelerating,han2017ese} vary from vary small to large. RNNFast shows from 4.7$\times$ to 64$\times$ speedup. Compared to \cite{Ferreira_16ReConfig} RNNFast has 19$\times$ less energy consumption.

Recently Fowers et al.\cite{brainwave_isca18} introduced Brainwave, an FPGA-based accelerator for RNN with no batching for real time AI. 
While a very efficient design, Brainwave has 50-70\% higher energy energy than RNNFast. Brainwave also shows poorer performance for smaller networks, but slightly better performance for large ones, compared to RNNFast. Note that this is not a quantitative apples-to-apples comparison to our design given that Brainwave uses 8 bit precision (vs 16 bit for RNNFast) and a 14nm techology node (vs. 32nm for RNNFast). 

\begin{table}[h]
\small
\centering
\scalebox{0.7}{
\begin{tabular}{|l|l|l|l|l|l|l|}
\hline
FPGA  & \multirow{2}{*}{Net size} & \multirow{2}{*}{Timesteps} &\multirow{2}{*}{run time($\mu$s)}  & \multirow{2}{*}{energy ($\mu$J)} & RNNFast  &   RNNFast  \\
 Design   &  &  &   &   &run time ($\mu$s) & energy ($\mu$J)\\
\hline
\cite{Ferreira_16ReConfig} & 32 & 1 & 1.586 &  0.8 & 0.332& 0.0419\\
\hline
\cite{mealey2018accelerating}&256&7735&42.48E3&NA&2.13E3&1.28E3\\
\hline
\cite{han2017ese}&1024& 1 &82.7& NA&1.29&12.8\\
\hline
\cite{brainwave_isca18} & 256-1k-2K&150-25-25 &425-74-74 & Est.: 425-1091-4356 & 117-58-110.7 & 252-643-2575\\
\hline
\end{tabular}
}

\caption{Energy and run time for FPGA-based RNNs.}
\label{tab:FPGA}
\vspace{-1em}
\end{table}
The Google TPU is also capable of running RNN workloads efficiently. In \cite{TPU_isca_17} they report up to 8$\times$ better performance for LSTM workloads compared to NVIDIA K80. RNNFast is up to $260\times$ faster than the newer NVIDIA P100 for workloads of similar size. 

\section{Other Related Work}
\label{sec:related}

Many customized accelerators for machines learning algorithms and DNNs have been proposed recently
\cite{chen_micro2014,chen_asplos2014,du_isca2015, 
liu_asplos2015, du_isca2015, han_isca2016, chen_isca2016, 
shafiee_isca2016, chi_isca2016, kim_isca2016, albericio_isca2016, 
likamwa_isca2016, reagenisca2016, liu_isca2016,chung_islped2016}. 
The majority of this work focuses on improving the performance of CNNs,
exploring the potential for resources sharing, leveraging emerging
memory technologies, optimizing basic operations, and developing domain specific methods.

\cite{han_isca2016} used compression of the network model to
reduce the memory footprint and accelerate real-time networks in which
batching cannot be employed to improve data re-use. Eyeriss
\cite{chen_isca2016} explored local data reuse of filter weights and activations in high-dimensional convolutions in order to minimize the energy of data movement.

Emerging memory technologies and in-memory processing have been
leveraged for CNN designs to address memory latency limitations and to
implement custom logic.  PRIME \cite{chi_isca2016} combined processor-in-memory
architecture and ReRAM-based neural network computation. The crossbar array
structure in ReRAM can be used to perform matrix-vector multiplication
as well as regular memory to increase memory space. PUMA \cite{ankit2019puma}, a recently proposed general-purpose and ISA-programmable accelerator built with ReRAM. It has a spatial architecture organized in cores, tiles, and nodes. PUMA features a microarchitecture, ISA, and compiler co-designed to optimize data movement and maximize energy and area efficiency. The PUMA design is more general than ISAAC~\cite{shafiee_isca2016}, and, as a result, it generally performs a bit worse in terms of throughput and energy efficiency. 
ReRAM-based DNN accelerators benefit from the speed and efficiency of the memristor crossbar; however the need for additional peripheral circuits such as ADCs and DACs, and other components, reduce the benefits of crossbar-based computation.

Neurocube \cite{kim_isca2016} proposed a programmable and scalable
digital neuromorphic architecture based on 3D high-density memory
integrated with a logic tier for efficient neural computing.  The design in \cite{Long_IJCNN_16} also used ReRAM cross bar for RNN acceleration for a case of human activity detection with small network size of 100 and simple vanilla RNN. 
Cambricon \cite{liu_isca2016} propose a novel domain-specific 
Instruction Set Architecture (ISA) for neural network accelerators. 
PuDianNao \cite{liu_asplos2015} focuses on a range of popular machine learning algorithms. 
However all these optimizations are CNNs/DNNs specific.
Chung et. al \cite{chung_islped2016} used DWM for CNN computations as well. They proposed a new design that replaces the ReRAM cross bar with a DWM-based CNN layer for dot product. However, they still use costly ADC/DAC circuits and also did not address DWM shift errors in their design.

\section{Conclusion}
\label{sec:conclusions}

The unprecedented growth of available data is accelerating the
adoption of deep learning across a wide range of applications
including speech recognition, machine translation, and language
modeling.  In this study, we propose RNNFast, a novel accelerator
designed for recurrent neural networks.  Our approach demonstrates that using domain wall memory is not only feasible, but also very efficient. We compare our design with a state-of-the-art  P100 NVIDIA GPU and find $21.8\times$ better performance with $70\times$ lower energy.

\bibliographystyle{ACM-Reference-Format}
\bibliography{rnn}

\end{document}